\documentclass[10pt,journal,compsoc]{IEEEtran}
\usepackage{graphicx}
\usepackage{subcaption}
\usepackage{enumitem}
\usepackage{hyperref} %
\hypersetup{colorlinks=true,linkcolor=blue,anchorcolor=blue,citecolor=blue}
\usepackage{multirow}
\usepackage{setspace}
\usepackage{multicol}
\usepackage{stfloats}
\newcommand{\mathbfz}{\mathbf{z}}

\usepackage[table]{xcolor}
\definecolor{lightgray}{RGB}{220, 220, 220} %
\usepackage{xspace}
\usepackage{bbding}
\usepackage{pifont}
\usepackage{ragged2e}
\newcommand{\etal}{\emph{et al.}\xspace}

\newcommand{\myblue}[1]{{\color{black}#1}}
\newcommand{\myred}[1]{{\color{black}#1}}
\definecolor{mygreen_rgb}{RGB}{0,1,0}
\newcommand{\mygreen}[1]{{\color[rgb]{0,0.5,0}\textbf{#1}}}
\usepackage{algorithm}
\usepackage{algorithmic}
\usepackage{amsmath}

\usepackage{amssymb}
\usepackage{booktabs}
\usepackage{multirow}

\begin{document}

\title{\huge MADiff: Motion-Aware Mamba Diffusion Models for Hand Trajectory Prediction on Egocentric Videos}%

\author{Junyi~Ma$^{\dag}$, Xieyuanli~Chen$^{\dag}$, Wentao~Bao, Jingyi~Xu, Hesheng~Wang$^{*}$
\thanks{\quad This work was supported by National Key R\&D Program of China (Grant No.2024YFB4708900), and Shanghai
Municipal Science and Technology Major Project (Grant No.2025SHZDZX025G01). It was also supported in part by the Natural Science Foundation of China under Grants 62225309, U24A20278, 62361166632, and U21A20480. Junyi~Ma and Hesheng~Wang are with IRMV Lab, School of Automation and Intelligent Sensing, Shanghai Jiao Tong University and State Key Laboratory of Avionics Integration and Aviation System-of-Systems Synthesis, Shanghai Key Laboratory of Navigation and Location Based Services, Shanghai 200240, China. Xieyuanli~Chen is with the College of Intelligence Science and Technology, National University of Defense Technology, Changsha 410073, China. Jingyi~Xu is with the Department of Electronic Engineering, Shanghai Jiao Tong University, Shanghai 200240, China.}
\thanks{\quad $^{\dag}$Equal contribution}
\thanks{\quad $^{*}$Corresponding author email: wanghesheng@sjtu.edu.cn}
}

\IEEEtitleabstractindextext{
\begin{abstract}
Understanding human intentions and actions through egocentric videos is important on the path to embodied artificial intelligence. As a branch of egocentric vision techniques, hand trajectory prediction plays a vital role in comprehending human motion patterns, benefiting downstream tasks in extended reality and robot manipulation. However, capturing high-level human intentions consistent with reasonable temporal causality is challenging when only egocentric videos are available. This difficulty is exacerbated under camera egomotion interference and the absence of affordance labels to explicitly guide the optimization of hand waypoint distribution.
In this work, we propose a novel hand trajectory prediction method dubbed MADiff, which forecasts future hand waypoints with diffusion models. The devised denoising operation in the latent space is achieved by our proposed motion-aware Mamba, where the camera wearer's egomotion is integrated to achieve motion-driven selective scan (MDSS). To discern the relationship between hands and scenarios without explicit affordance supervision, we leverage a foundation model that fuses visual and language features to capture high-level semantics from video clips. \myblue{Comprehensive experiments conducted on five public datasets with the existing and our new evaluation metrics demonstrate that MADiff predicts comparably reasonable hand trajectories compared to the state-of-the-art baselines.} We have released our code and pretrained models of MADiff at the project page: \url{https://irmvlab.github.io/madiff.github.io}.
\end{abstract}


}

\maketitle
\IEEEdisplaynontitleabstractindextext
\IEEEpeerreviewmaketitle

\section{Introduction}
\label{sec:intro}

Embodied artificial intelligence requires deep comprehension of human behaviors and flexible techniques, transferring general skills from daily human activities to robotics. Extracting reusable and transferable knowledge from internet-scale human videos is regarded as an efficient way to understand human intentions and actions. Many efforts have been made to achieve action recognition and anticipation \cite{xu2023dynamic,zheng2020dynamic,zhang2024object,zheng2023icme,qi2024uncertainty,bao2021evidential,bao2021drive}, temporal action localization \cite{wang2024temporal,chen2023uncertainty,li2024detal,bao2022opental}, gaze prediction \cite{zhang2017deep,lai2024eye,li2013learning,li2018eye}, hand trajectory prediction \cite{liu2022joint,bao2023uncertainty,liu2020forecasting,shan2020understanding,ma2024diff}, object affordance extraction \cite{liu2022joint,ye2023affordance,liu2020forecasting,xu2023interdiff,ma2024diff}, and object interaction anticipation \cite{Pasca_2024_CVPR,furnari2017next,bertasius2016first,ragusa2023stillfast}. Among them, hand trajectory prediction (HTP) is a comparably challenging task that aims to anticipate how humans will behave in the near future, moving beyond just estimating action categories or gaze direction. This task is valuable for collecting offline data, predefining the action space for robot learning, and assisting human activities in extended reality applications \cite{bahl2023affordances,li2022egocentric,bao2023uncertainty}.

\begin{figure}[t]
  \centering
  \includegraphics[width=1\linewidth]{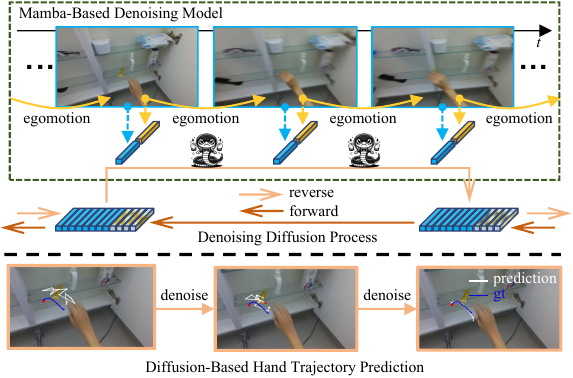}
  \caption{MADiff reconstructs future latents conditioned on past latents in the diffusion process. A Mamba-based model is designed to achieve motion-driven selective scan in the denoising process. The reconstructed future latent features are utilized to generate hand trajectory predictions.}
  \label{fig:motivation}
  \vspace{-0.5cm}
\end{figure}

Considering that humans use egocentric vision to perceive the world and guide daily tasks, several notable convolution- and transformer-based HTP approaches \cite{liu2020forecasting,liu2022joint,bao2023uncertainty,ma2024diff} have been proposed in recent years to forecast incoming hand positions with only egocentric videos as inputs. 
Despite achieving acceptable prediction results, several challenging problems remain to be solved:
\begin{itemize}[leftmargin=1em]
    \item Camera egomotion guidance has not been seamlessly integrated into the state transition of the HTP process to narrow the motion-related gaps we discovered: 1) Predicting the 3D trajectories of future hand movements directly projected onto the 2D egocentric image plane, presents a challenging problem due to spatial ambiguities. There exists a noticeable disparity between the movements observed in 2D pixels and the corresponding 3D physical actions, which can be mitigated by camera egomotion. 2) With the past egocentric video as input, we predict future hand waypoints on a predefined ``canvas'' such as the image plane of the first observation. However, the past hand positions and scene information within the other frames are observed in different views with respect to the canvas view due to the existence of camera egomotion. \myred{This problem is inevitable when HTP is deployed on downstream applications like mixed reality. When humans wear VR headsets for hand-object interaction, it is impractical to keep their heads perfectly still, resulting in persistent camera egomotion.}
    \item HTP models are often optimized along with ground-truth (GT) object affordances besides hand waypoints \cite{liu2020forecasting,liu2022joint,ma2024diff}. This respects the fact that hand trajectories typically interact with active objects based on human intentions as an oracle. Understanding hand movements involves being aware of both hand positions and environmental situations concurrently. However, annotating object affordances is labor-intensive \cite{liu2022joint,mur2024aff} compared to labeling hand trajectories. There is no off-the-shelf detector that can automatically and accurately identify the active objects interacted with a hand trajectory, attaining the quality of producing ground-truth.
    The previous work~\cite{darkhalil2022epic} shows that the performance of the existing detectors varies significantly across the two tasks, next active object detection and hand detection. Therefore, ground-truth object affordances are not always available due to a lack of manual labeling and low-quality automatic annotation.
    In the absence of object affordance labels to aid optimization, the inner correlation between hand motion and semantics in observations is hard to extract in a manner that aligns with human intentions by HTP models. 
    \item Causality and motion continuity constraints are often overlooked in the context of using trendy convolution or transformer supervised by waypoint displacement. Temporal causality is inherent in both hand motion and its parallel camera wearer's egomotion changes, as the hand and body are simultaneously guided by high-level intentions and the movement patterns of the hand are closely linked to those of the body. However, convolution- and transformer-based models \cite{liu2020forecasting,liu2022joint,bao2023uncertainty,ma2024diff} suffer from modeling the state transition by unexplainable attention mechanisms, and fail to selectively capture temporal causality considering the two entangled movement patterns. Moreover, the existing loss functions for constraining trajectory prediction are insufficient to adequately determine the optimization direction in line with the potential physical model of hand movements. \myred{We also argue that harnessing causality and motion continuity constraints is a critical step toward broader downstream applications like video-based human-robot skill transfer.}
\end{itemize}
To address these existing gaps, we propose MADiff, a motion-aware Mamba diffusion model to predict future hand waypoints on egocentric videos. To overcome the challenge of observation semantics caused by a lack of object affordances, we first exploit a foundation model in MADiff to fuse visual and language features in a generalizable manner, thereby capturing high-level semantics from 2D input images without the need for affordance labels. We demonstrate that using a visual grounding model with text guidance as the backbone to generate task-related features from observations significantly enhances hand trajectory prediction, compared to models that are task-agnostic or trained from scratch. Subsequently, we convert both semantic features and past trajectory features to sequential latents. Inspired by the strong generative capability of diffusion models \cite{ho2020denoising,song2020denoising} in predictive tasks \cite{yu2023video, rasul2021autoregressive,li2022generative}, we implement denoising diffusion within the above-mentioned latent space, using the devised Mamba model with motion-driven selective scan (MDSS) to recover future latents conditioned on past sequential features as shown in Fig.~\ref{fig:motivation}. These reconstructed latents are then transformed into the final predicted hand waypoints. Here, we extend the selective state space models with scan computation (S6)~\cite{gu2023mamba} by incorporating the camera wearer's egomotion (camera homography) to achieve motion-driven state transition. This helps to fill the motion-related gaps caused by different prediction canvas and 2D-3D aliasing, and enhances the explainability in temporal causality of the entangled
movement patterns. We additionally design a continuous-discrete-continuous (CDC) operation for denoising diffusion combining the strengths of autoregressive (AR) models and iterative non-autoregressive (iter-NAR) models. Furthermore, we propose an effective angle/length supervision strategy for the training paradigm to improve the directionality and stability of predicted hand trajectories. This overcomes the challenge of optimizing HTP models with motion continuity constraints.

The main contributions of this paper are fourfold:
\begin{itemize}[leftmargin=1em]
\item We propose MADiff, the pioneering diffusion-based method for predicting hand trajectories, featuring a devised motion-aware Mamba as the denoising model. A novel motion-driven selective scan pattern is tailored to facilitate a suitable state transition in Mamba-based denoising, comprehensively considering both hand motion and camera egomotion patterns to capture temporal causality. Moreover, MADiff bridges autoregressive models and iterative non-autoregressive models, building a novel generative paradigm for hand trajectory prediction.
\item We first propose using the fusion of visual and language prompts for semantics extraction on 2D video clips in the realm of hand trajectory prediction. This addresses the challenge of high-level scene understanding due to the absence of affordance labels. Besides, the consistency inherent in deep semantic features also naturally aligns with human intention consistency. By seamlessly integrating the multimodal cues, we lay the foundation for a new scheme of semantic richness in hand trajectory prediction.
\item We first emphasize the importance of directionality and stability in the field of hand trajectory prediction. We accordingly design new loss functions for optimization implicitly constrained by physical models of hand motion, leading to more plausible prediction results. 
\item We conduct comprehensive experiments based on the existing and our new evaluation metrics to demonstrate that MADiff predicts comparably reasonable hand trajectories compared to state-of-the-art baselines. \myred{We also experimentally demonstrate MADiff's potential to provide flexible HTP solutions tailored to specific action verbs.}
\end{itemize}
This paper is organized as follows. Sec.~\ref{sec:related_work} reviews the related works in egocentric vision and some cutting-edge techniques in diffusion models and Mamba. Sec.~\ref{sec:preliminaries} introduces the preliminaries of our work. Sec.~\ref{sec:MADiff} details the design of our proposed MADiff. Sec.~\ref{sec:exp} showcases the experimental results quantitatively and qualitatively. Finally, Sec.~\ref{sec:conclusion} concludes the paper and provides our insights.

\section{Related Work}
\label{sec:related_work}
\subsection{Understanding Hand-Object Interaction}

Hand-object interaction (HOI) comprehension helps guide the downstream tasks in computer vision and robot systems.
In the early stage, Calway \etal \cite{calway2015discovering} build connections between human tasks and corresponding objects, which highlights an object-centric comprehension across diverse interaction modes. In contrast, Liu \etal \cite{Liu_2017_ICCV} emphasize capturing dynamic attributes of objects, underscoring the relationship between object-centric interactions and goal-directed activities. After that, more and more works contribute to HOI understanding by pixel-wise semantic segmentation \cite{schroder2017hand,darkhalil2022epic,zhang2022fine,Higgins_2023_CVPR}, bounding-box-wise detection \cite{furnari2020rolling,fan2021understanding,Shiota_2024_WACV,shan2020understanding}, fine-grained hand/object pose estimation \cite{romero2022embodied,zhou2020monocular,Lin_2023_CVPR,yang2022artiboost,liu2021semi,lim2013parsing}, and contact field estimation \cite{yang2024learning,yang2021cpf}. Ego4D \cite{grauman2022ego4d} further conducts a standard benchmark that evaluates HOI understanding based on predefined subtasks. However, only comprehending what has happened to humans and environments (objects) is not enough in many applications, where future possible hand positions or object states are required to plan downstream tasks.

\vspace{-0.2cm}

\subsection{Predicting Future Hand Trajectories}
\label{sec:htp_related_work}

Given sequential egocentric observations, forecasting future hand positions is a valid approach extended in time horizons to understanding human actions and intentions in AR/VR and robot manipulation. Although it is technically possible to predict fine-grained hand keypoints by extending hand keypoint estimation \cite{mucha2024my,Mueller_2017_ICCV,garcia2018first}, directly forecasting 2D hand waypoints in the near future focuses more on understanding high-level human intentions, which avoids large error accumulation and benefits running efficiency compared to predicting multiple complicated keypoints.
FHOI~\cite{liu2020forecasting} samples future hand waypoints through motor attention following a 3D convolutional network, using stochastic units to model the uncertainty. Following its task definition, the object-centric transformer (OCT) \cite{liu2022joint} is further proposed combined with conditional variational autoencoders \cite{sohn2015learning}. VRB \cite{bahl2023affordances} designs an affordance model to simultaneously predict contact point heatmap and post-contact hand trajectories. To additionally capture the uncertainty of predicted trajectories, an uncertainty-aware state space transformer (USST) \cite{bao2023uncertainty} is proposed to model the state transition in the unrolling process. More recently, Diff-IP2D \cite{ma2024diff} builds a new diffusion-based paradigm for hand-object interaction. 
Although it attempts to mitigate the negative effect of camera motion, its denoising process with motion feature integration does not follow the specific hand state transition process, leading to a weak awareness of causality in HTP. In contrast, in this work, we propose a motion-aware Mamba with a motion-driven selective scan to achieve a more reasonable denoising process. Moreover, most existing HTP approaches \cite{liu2020forecasting,liu2022joint,bahl2023affordances,ma2024diff} need affordance labels such as object contact points to guide the optimization of hand waypoint distribution. We avoid the redundancy requirement by utilizing a foundation model to semantically comprehend the relationships between hands and scenarios. 

\vspace{-0.1cm}
\subsection{Generative Paradigm in Egocentric Vision}
Generative models have been demonstrated to perform well across multiple subfields of egocentric vision. EgoGAN~\cite{jia2022generative} utilizes a Generative Adversarial Network (GAN) to forecast future hand masks conditioned on encoded video representation and predicted future head motion. Zhang \etal \cite{zhang2017deep} also use GAN-based model to generate future frames and predict their temporal saliency maps which reveal the probability of gaze locations.
With the advent of diffusion models~\cite{ho2020denoising,song2020denoising}, diffusion-based generative modeling generally beats discriminative and GAN-based modeling in egocentric vision, including egocentric video prediction \cite{luo2024put,chang2023look}, human mesh recovery~\cite{zhang2023probabilistic,liu2023egohmr}, 3D HOI reconstruction \cite{Ye_2023_ICCV,zhu2023get}, and 3D HOI synthesizing \cite{zhang2024hoidiffusion,ye2023affordance}. Zhong \etal \cite{zhong2023diffant} propose a diffusion-based method namely DiffAnt for long-term action anticipation. It follows the query-based scheme \cite{carion2020end,zhu2020deformable} for decoding future embeddings to action labels. Li \etal \cite{li2023ego} utilize a diffusion model conditioned on the estimated head pose to infer the full-body pose.
In this work, we also propose a diffusion-based generative paradigm for HTP on egocentric videos, combined with the devised Mamba as the denoising model. 

\vspace{-0.3cm}

\subsection{Mamba in Time Series Forecasting}
As a trendy state space model (SSM), Mamba \cite{gu2023mamba} exhibits competitive ability in modeling long-range dependency and improving computational efficiency compared to transformer \cite{vaswani2017attention}. It is built upon a selection mechanism and thus has a context-aware ability to compress and propagate effective information in the state transition process. Moreover, Mamba also uses a hardware-aware algorithm for the parallel associative scan. Recently, some Mamba-based methods for time series forecasting have been proposed. For example, SiMBA \cite{patro2024simba} uses EinFFT for channel modeling and Mamba for token mixing, presenting solid performance on multivariate long-term forecasting tasks. TimeMachine \cite{ahamed2024timemachine} combines an inner Mamba and an outer Mamba to address channel-mixing and channel-independence problems simultaneously while selecting global and local contexts at multiple scales.
S-Mamba \cite{wang2024mamba} and Bi-Mamba+ \cite{liang2024bi} both consider the bidirectional scan pattern on sequential tokens, breaking the limitation of incorporating antecedent variates. 

Compared to these time series forecasting methods designed task-agnostically, in this work, we focus on the specific realm of hand trajectory prediction and develop a novel motion-aware Mamba regarding the characteristics of the hand movements and camera wearer's egomotion. Moreover, we integrate the devised Mamba blocks into a diffusion process, which builds a novel paradigm bridging autoregressive and iterative
non-autoregressive models, and provides a basic framework for time series forecasting.
Our experiments show that our motion-driven selective scan (MDSS) performs better than the recent bidirectional scan pattern \cite{wang2024mamba,liang2024bi} for hand trajectory prediction due to the unreasonable inversion of causality and human motion pattern inherent in the bidirectional mechanism (see Sec.~\ref{sec:abla_on_mdss}).

\begin{figure*}[t]
  \centering
  \includegraphics[width=0.87\linewidth]{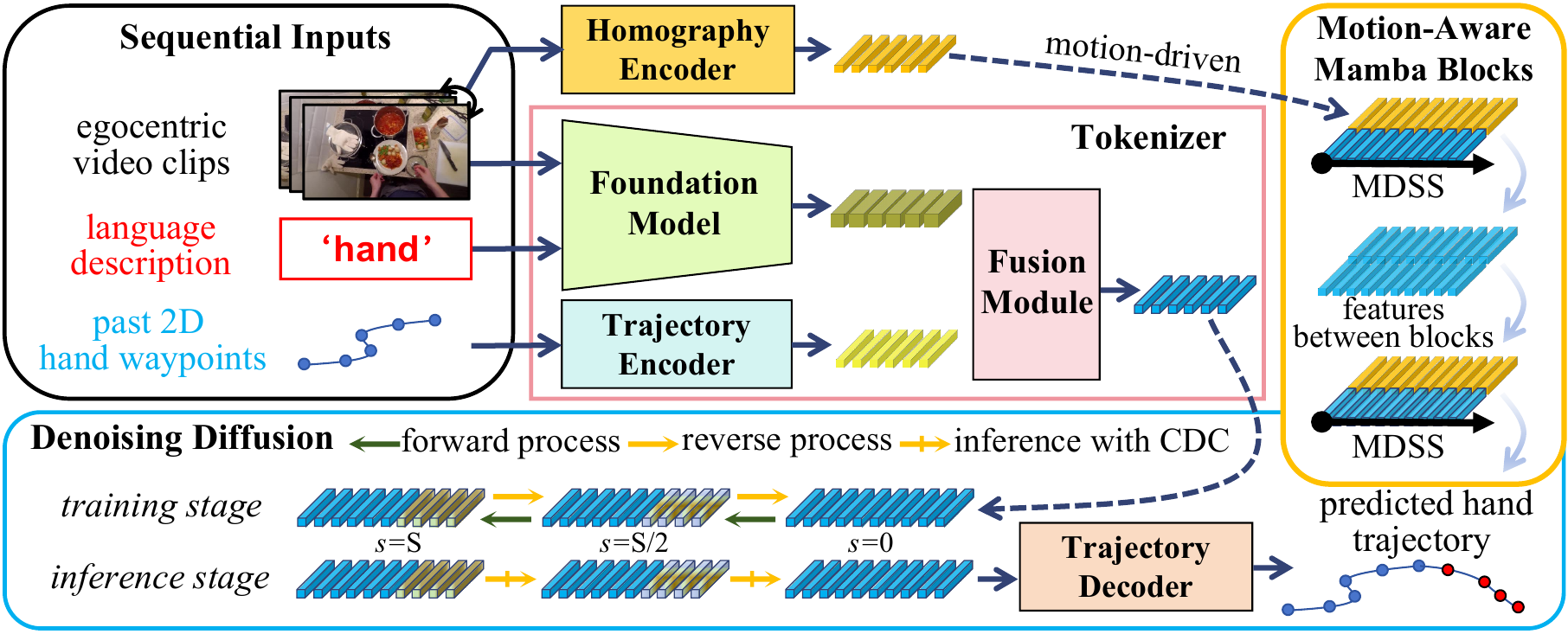}
  \caption{\myred{System overview of MADiff. We use egocentric video clips, language description, and past 2D hand waypoints as inputs and design a Mamba-based and motion-driven denoising diffusion process to predict future 2D hand trajectories.}}
  \label{fig:MADiff_arch}
  \vspace{-0.5cm}
\end{figure*}

\section{Preliminaries}
\label{sec:preliminaries}

\subsection{Task Definition}
\label{sec:task_definition}

Given the video clip of past egocentric observations $\mathcal{I}=\{I_t\}_{t=-N_\text{p}+1}^{0}$ and sequential past 2D hand waypoints $\mathcal{H}^\text{p}=\{H_t\}_{t=-N_\text{p}+1}^{0} (H_t \in \mathbb{R}^{2})$, our objective is to predict future hand trajectories $\mathcal{H}^\text{f}=\{H_t\}_{t=1}^{N_\text{f}} (H_t \in \mathbb{R}^{2})$, where $N_\text{p}$ and $N_\text{f}$ correspond to the number of frames in the past and future time horizons. It can be represented by modeling an unknown joint distribution of future hand waypoints $p_{\Phi}(\mathcal{H}^\text{f}|\mathcal{H}^\text{p},\Theta)$ where $\Phi$ denotes a predictive model and $\Theta$ encompasses additional conditions.
Following the previous works \cite{bao2023uncertainty,liu2022joint}, we predict the future positions of both hands on a fixed image plane of the input videos, e.g., the first observed image as the prediction canvas.
\myblue{Here, we only focus on the 2D predictive task, since past 3D hand trajectories are not always available due to limited onboard sensors. In contrast, 2D hand trajectories can be efficiently extracted using off-the-shelf hand detectors~\cite{shan2020understanding}. Besides, we argue that internet-scale 2D egocentric video data is more widely accessible than 3D data and is more likely to serve as a shortcut for achieving embodied intelligence. Many downstream applications such as screen overlays for immersive interaction of extended reality and human-robot skill transfer oftentimes operate in 2D spaces~\cite{bahl2023affordances,chang2023look}, where accurate 2D prediction is both sufficient and practical.}

\vspace{-0.3cm}

\subsection{Diffusion Models}
\label{sec:diffusion_models}

The diffusion models \cite{ho2020denoising,song2020denoising} can progressively corrupt the inputs into noisy features and subsequently recover them based on a devised denoising model. 
Here we use its generative capability for predicting future hand trajectories on 2D egocentric videos. We argue that diffusion models can well model highly dynamic patterns inherent in complex distributions of future hand motion. Besides, the HTP iteration limited in the time axis can be extended to a more flexible diffusion denoising process. 
Initially, we map the input images and past hand waypoints into a latent space, denoted as $\mathbfz_0 \sim q(\mathbfz_0)$. This latent representation is then corrupted into standard Gaussian noise, represented as $\mathbfz_S \sim \mathcal{N}(0, \mathbf{I})$. During the forward process, the perturbation operation is described by $q(\mathbfz_s|\mathbfz_{s-1}) = \mathcal{N}(\mathbfz_s; \sqrt{1-\beta_s}\mathbfz_{s-1}, \beta_s\mathbf{I})$, where $\beta_s$ is the predefined variance scales. In the reverse process, we employ a denoising diffusion model to gradually reconstruct the latent representation $\mathbfz_0$ from the noisy $\mathbfz_S$. The denoised features are then transformed into the predicted future hand trajectories. In this work, we will elaborate on solving the problems of generating reasonable latents, building a novel task-related denoising model, integrating effective denoising guidance, and designing suitable training and inference schemes for diffusion models in the hand trajectory prediction task.

\vspace{-0.3cm}
\subsection{State Space Models of Mamba}
\label{sec:mamba}

State space model (SSM) of Mamba~\cite{gu2023mamba}, built upon a selection mechanism, has a context-aware ability to compress and propagate effective information in the state transition. It utilizes first-order differential equations to link input and output sequences via hidden states.
Our approach utilizes the discrete version of the continuous-time SSM in Mamba:
\begin{align}
\bar{A} &= e^{\Delta A}, \\ 
\bar{B} &= (e^{\Delta A} - I) A^{-1} B,  \label{eq:mamba_ori0} \\ 
h_k &= \bar{A} h_{k-1} + \bar{B} x_k, \label{eq:mamba_ori1} \\
y_k &= C h_k , 
\label{eq:mamba_ori}
\end{align}
where $A$ serves as the evolution parameter, $B$ and $C$ act as projection parameters, and $\Delta$ is a timescale parameter for the discretization. The structured state space model (S4)~\cite{gu2021efficiently} initializes $A$ 
by HIPPO theory~\cite{gu2020hippo}. Mamba further extends S4 to S6 by forcing $B$, $C$, and $\Delta$ to be functions of the input. 
In this work, we propose naturally utilizing the camera wearer's egomotion information ($m_{t-1}\xrightarrow{} m_{t}$), i.e., homography egomotion features, to drive the state transition process ($h_{t-1}\xrightarrow{} h_{t}$) in Mamba, and seamlessly integrate the state space model into a denoising diffusion process, bridging autoregressive and iterative non-autoregressive schemes in the hand trajectory prediction task.

\vspace{-0.5cm}

\section{Proposed Method}
\label{sec:MADiff}

\subsection{System Overview}
\label{sec:MADiff_archtecture}

\myblue{The overall pipeline of MADiff is illustrated in Fig.~\ref{fig:MADiff_arch}.} \myred{The inputs for MADiff encompass past sequential egocentric images and 2D hand waypoints within the given video clip, as well as the language description as the proposed text prompt. Tokenizer first generates visual-language features through a foundation model, encodes past hand waypoints to sequential intermediate features with the trajectory encoder, and then fuses them by the fusion module (Sec.~\ref{sec:tokenizer}). The output of the tokenizer is the tokenized latents utilized by our proposed motion-aware Mamba (Sec.~\ref{sec:motion-mamba}) in the devised Mamba-based denoising diffusion model (Sec.~\ref{sec:mamba_diff}), where we design a motion-driven selective scan to recover the future latents conditioned on the past latents. Ultimately, the trajectory decoder transforms the reconstructed latent features to predicted future hands waypoints. We design new training loss functions and inference operations for MADiff, which can be found in Sec.~\ref{sec:train_ref}.}

\vspace{-0.3cm}

\subsection{Tokenizer}
\label{sec:tokenizer}

\myblue{The devised tokenizer of MADiff contains a foundation model, a trajectory encoder, and a fusion module.} It exploits three types of input data: past egocentric video clips, language descriptions, and past 2D hand waypoints. \myblue{We fuse multimodal cues to represent the observation at each timestamp by the tokenizer and enhance the prediction performance of MADiff, which can also serve as the foundation for a new scheme of semantic richness in the HTP literature.}

\begin{figure}[t]
  \centering
  \includegraphics[width=0.85\linewidth]{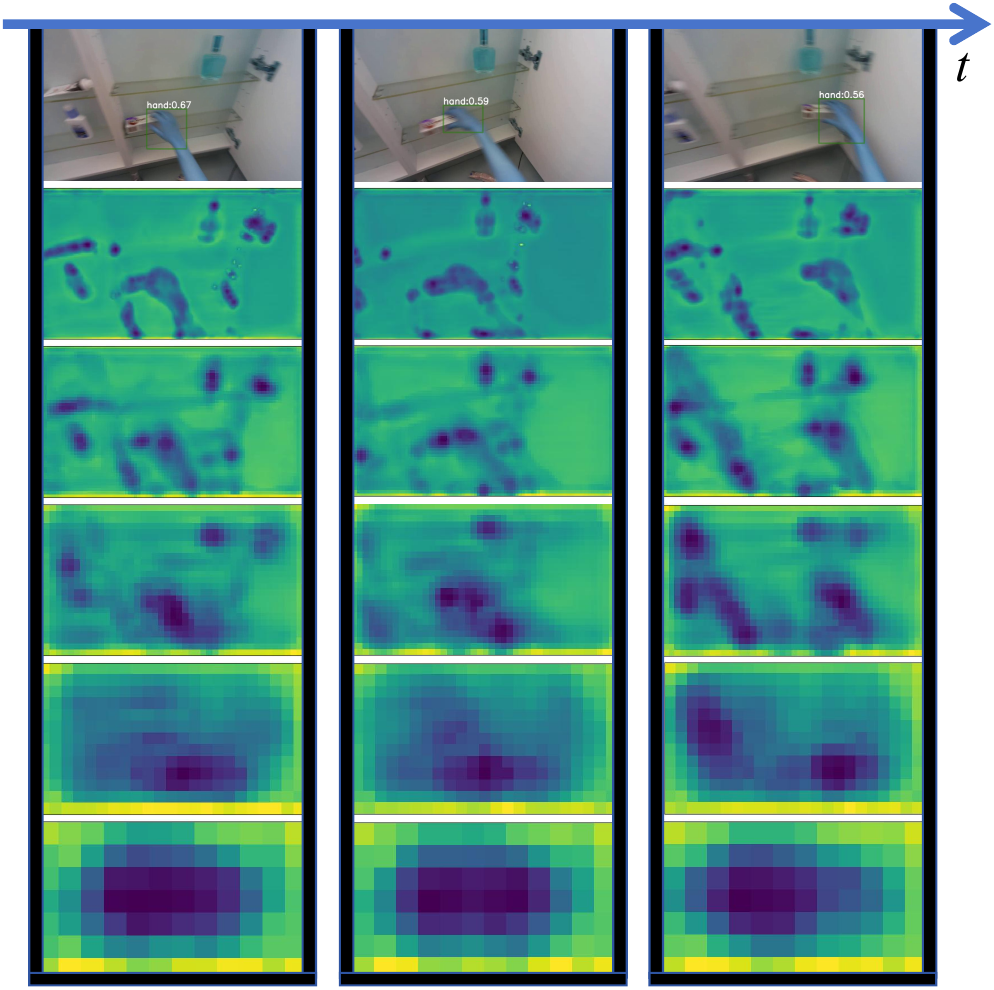}
  \caption{Visual-language fusion features extracted from a video example of EgoPAT3D-DT \cite{li2022egocentric,bao2023uncertainty} dataset by GLIP (average pooling over the channel dimension). GLIP attends to the target $\mathtt{hand}$ of text prompt and possible active objects, therefore extracting semantics with no need for affordance supervision. The deepest features align with the consistency in human intention, and therefore can be regarded as a high-level understanding of the interaction process. The sizes of the example feature maps from top to bottom (from shallow to deep in GLIP deep fusion) are $256\times 100\times 180$, $256\times 50\times 90$, $256\times 25\times 45$, $256\times 13\times 23$, and $256\times 7\times 12$.}
  \label{fig:glip_feats}
  \vspace{-0.3cm}
\end{figure}

\textbf{Foundation Model:} \myblue{MADiff exploits a powerful foundation model, the widely-used GLIP \cite{Li_2022_CVPR} to generate visual-language fusion features from sequential past observations (as shown in Fig.~\ref{fig:glip_feats}).} In contrast to existing works~\cite{bao2023uncertainty,liu2020forecasting,liu2022joint, ma2024diff} only using visual inputs, we additionally consider the text prompt $\mathtt{hand}$ when MADiff captures past environment observations and predicts future hand states. The visual grounding ability of GLIP enables our MADiff to semi-implicitly capture hand poses and hand-scenario relationships within each 2D image frame. This guides the optimization of hand waypoint distribution, demonstrated in Sec.~\ref{sec:abla_on_inputs}, without the need for affordance supervision required by previous works~\cite{liu2020forecasting,liu2022joint, ma2024diff,bahl2023affordances}. 
We also discovered that the deepest features averaged over the channel dimension at continuous timestamps exhibit potential consistency, shown in Fig.~\ref{fig:glip_feats}, which aligns with the consistency in human intention during the interaction process.
The joint application of the foundation model and language description enhances MADiff's generalization ability and deployment efficiency compared to those using backbones trained on specified HOI datasets from scratch~\cite{liu2022joint,ma2024diff}, and concurrently holds HTP task specificity in contrast to those using off-the-shelf pretrained backbones~\cite{bao2023uncertainty,liu2020forecasting}. Specifically, we extract the outputs of the deepest cross-modality multi-head attention module (X-MHA) in GLIP, which are denoted as the semantic features $\mathcal{X}^{\text{sem}}=\{X^{\text{sem}}_t\}_{t=-N_\text{p}+1}^{L}$ for hand trajectory prediction. $L$ equals $N_\text{f}$ during training and is set to $0$ during inference since future observations are unavailable in real deployment and are replaced by sampled noise in the subsequent diffusion models.

\begin{figure}[t]
  \centering
  \includegraphics[width=0.9\linewidth]{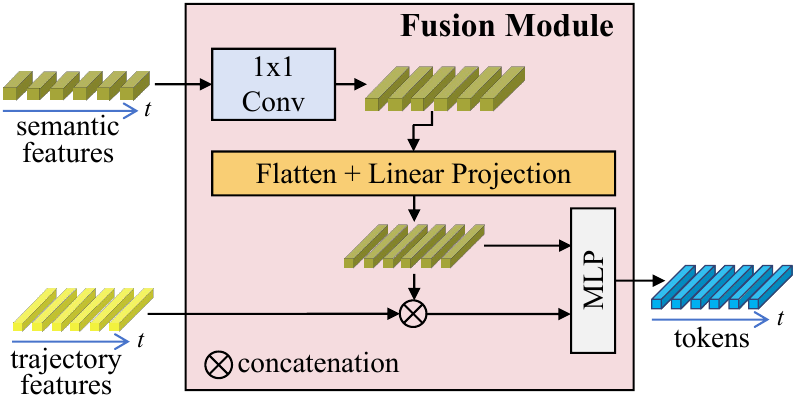}
  \caption{\myred{Architecture of the fusion module in MADiff. It fuses semantic features from the foundation model with trajectory features from the trajectory encoder to generate tokens for the following diffusion model.}}
  \label{fig:fusion_module}
  \vspace{-0.5cm}
\end{figure}

\textbf{Trajectory Encoder and Fusion Module:} We use multilayer perceptrons (MLPs) as the trajectory encoder, which converts the sequential 2D hand waypoints $\mathcal{H}=\{H_t\}_{t=-N_\text{p}+1}^{L}$ to intermediate trajectory features $\mathcal{X}^{\text{traj}}=\{X^{\text{traj}}_t\}_{t=-N_\text{p}+1}^{L}$ in parallel. \myblue{The fusion module in Fig.~\ref{fig:fusion_module} first adopts $1\times 1$ convolution as well as a linear projection to adjust the spatial and channel dimensions of $\mathcal{X}^{\text{sem}}$ to match $\mathcal{X}^{\text{traj}}$, and subsequently uses MLP to fuse adapted $\mathcal{X}^{\text{sem}}$ and $\mathcal{X}^{\text{traj}}$ to $\mathcal{F}=\{F_t\}_{t=-N_\text{p}+1}^{L}$ as tokens for all timestamps $t$, also as latents for the following devised diffusion process.}

\subsection{Motion-Aware Mamba}
\label{sec:motion-mamba}

\begin{figure}[t]
  \centering
  \includegraphics[width=1\linewidth]{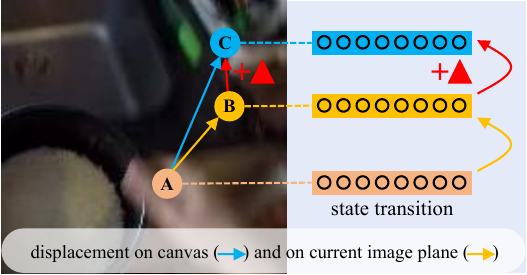}
  \caption{Start waypoint \textbf{A} and predicted end waypoint \textbf{B} are on the current image plane. Predicted waypoint \textbf{C} corresponds to the same 3D hand position as \textbf{B} but exists on the canvas image plane. The prediction model is empirically sensitive to the current displacement on (\textbf{A}$\xrightarrow{}$\textbf{B}), which needs to be shifted by an additional egomotion vector transformed from the homography matrix, to get the end waypoint C on canvas (\textbf{A}$\xrightarrow{}$\textbf{B}$\xrightarrow{}$\textbf{C}). We thus consider an additional feature update from the same homography matrix for state transition in the latent space intuitively analogous to the shift in the 2D image space, as Eq.~(\ref{eq:mdss1}) depicts.}
  \label{fig:intuition}
  \vspace{-0.5cm}
\end{figure}

MLP, convolutional layers, and transformers may struggle to capture temporal causality inherent in hand movements due to a lack of state transition with an explicit selective mechanism along the time axis. \myblue{MADiff instead integrates Mamba~\cite{gu2023mamba} into continuous denoising steps to selectively capture temporal causality.} Due to the inherent motion interference/gaps related to prediction canvas and 2D-3D aliasing mentioned in Sec.~\ref{sec:intro}, we further integrate egomotion features into Mamba's selective scan process, leading to the proposed motion-driven selective scan (MDSS):
\begin{align}
h_t &= \bar{A} h_{t-1} + \bar{B} [x_t^{\text{T}}, \textbf{0}]^{\text{T}} + \bar{B} [\textbf{0},m_t^{\text{T}}]^{\text{T}}, \label{eq:mdss1} \\ 
y_t &= C h_t ,
\label{eq:mdss2}
\end{align}
where $x_t$ denotes the $t\,\text{th}$ fusion tokens in the sequential latents. \myred{$m_t$ is the $t\,\text{th}$ egomotion feature transformed from the homography matrix between $t\,\text{th}$ frame and the canvas frame by the homography encoder.} To calculate the homography matrix, we first extract SIFT descriptors~\cite{lowe2004distinctive} to determine pixel correspondences between two consecutive images from previous observations. Subsequently, we compute the homography matrix using RANSAC \cite{fischler1981random} which seeks a transformation that maximizes the number of inliers among the keypoint pairs.
\myblue{As noted in Eq.~(\ref{eq:mdss1}), we introduce an additional term related to the homography feature $m_t$ to achieve a shift to the original state transition in Eq.~(\ref{eq:mamba_ori1}).} This operation corresponds to the intuition that the position of each hand waypoint projected to the fixed image plane (e.g., the one of the last observation) used as prediction canvas equals the position in its original image plane shifted by an additional displacement of egomotion homography, as shown in Fig.~\ref{fig:intuition}. It also implicitly transforms hand movement-related features into a more easily predictable latent space through egomotion features, analogous to predicting on the canvas image plane. We therefore concurrently consider the two entangled motion patterns, the hand motion pattern implicit in $h_t$ and the camera egomotion pattern implicit in $m_t$ during state transition following the fact that the hands and body move in a physically coordinated manner.
Eq.~(\ref{eq:mdss1}) can be further rewritten as:
\begin{align}
h_t &= \bar{A} h_{t-1} + \bar{B} [x_t, m_t],
\label{eq:mdss3} 
\end{align}
where we denote the concatenation of $x_t$ and $m_t$ along the channel dimension as $[x_t, m_t]$ for brevity. \myblue{We do not use the sum of $x_t$ and $m_t$ here because $\bar{B}$ can adaptively reweight the two features.} Besides, $B$ and $C$ in Eq.~(\ref{eq:mamba_ori0}) and Eq.~(\ref{eq:mamba_ori}) are also projection functions of the input $[x_t, m_t]$, and thus are also referred to as motion-aware projection matrices. The additional motion-related term in Eq.~(\ref{eq:mdss1}) and matrices $B$ and $C$ being functions of egomotion jointly determine the motion-driven property in our proposed selective scan pattern. Here, we do not let matrix $A$ be a function of egomotion, because it stably encapsulates historical information, solving long-range dependency inherent in sequential past egomotion and other fusion features following the HIPPO theory~\cite{gu2020hippo}.
Ultimately, the output signals can be computed in parallel by the discrete convolution of the input sequence:
\begin{align}
\bar{K} &= (C\bar{B}, C\overline{AB}, \ldots, CA^{N_\text{p}+N_\text{f}-1}\bar{B}), \label{eq:proj_mat}\\
{\color{black} y} &{\color{black}=} {\color{black}[\mathbf{x}, \mathbf{m}]*\bar{K},}
\label{eq:mdss_par}
\end{align}
where $N_\text{p}+N_\text{f}$ corresponds to the length of the holistic hand trajectory. \myblue{$\mathbf{x}$ and $\mathbf{m}$ denote the concatenation of all the fusion tokens $x_t$ and egomotion features $m_t$ along the temporal dimension, respectively.}
\myblue{It is worth noting that our proposed motion-aware Mamba avoids the quadratic cost of the attention mechanism in transformer and simultaneously exhibits reasonable explainability in the state transition of hand movements. The motion-driven selective scan also adaptively retains critical causality across temporal sequences and narrows the inherent gaps caused by egomotion, outperforming the static weights of MLPs with limited sequential dependency modeling capacity. 
}

\subsection{Mamba in Denoising Diffusion}
\label{sec:mamba_diff}

We seamlessly integrate our devised motion-aware Mamba block into the continuous denoising diffusion process.
\myblue{In each denoising step of MADiff, we utilize multiple stacked motion-aware Mamba blocks to recover future latents.} The forward process is only implemented during training and the reverse process is required for both the training and test pipeline, which will be extensively analyzed in Sec.~\ref{sec:train_ref}.

\textbf{Forward Process:} We implement partial noising~\cite{gong2022diffuseq} in the forward process during training.
The output of the fusion module is first extended by a Markov transition $q(\mathbfz_0|F_t)= \mathcal{N}(F_t,\beta_0\textbf{\text{I}})$, where $F_t \in \mathbb{R}^{(N_\text{p}+N_\text{f})\times a}$. In each following forward step of the diffusion model, we implement $q(\mathbfz_s|\mathbfz_{s-1})$ by adding noise to the future part of $\mathbfz_{s-1}$, i.e., $\mathbfz_{s-1}[N_\text{p}\!+\!1\!:\!N_\text{p}\!+\!N_\text{f}]$. 

\textbf{Reverse Process:} After $\mathbfz_S$ is derived after the forward process, our proposed motion-aware Mamba is exploited to denoise $\mathbfz_S$ to $\mathbfz_0$. \myblue{Considering the guidance of egomotion features $\mathbf{m}$, the reverse process can be modeled as $p_{\scriptscriptstyle{\text{Mamba}}}(\mathbfz_{0:S}):=p(\mathbfz_S)\prod_{s=1}^{S}p_{\scriptscriptstyle{\text{Mamba}}}(\mathbfz_{s-1}|\mathbfz_{s},\mathbf{m})$. Our $\ell$ stacked Mamba blocks $f_{\scriptscriptstyle\text{Mamba}}(\mathbfz_s,s,\mathbf{m})$ predicts the injected noise for each forward step with $p_{\scriptscriptstyle{\text{Mamba}}}(\mathbfz_{s-1}|\mathbfz_{s},\mathbf{m})=\mathcal{N}(\mathbfz_{s-1};\mu_{\scriptscriptstyle{\text{Mamba}}}(\mathbfz_s,s,\mathbf{m}),\sigma_{\scriptscriptstyle{\text{Mamba}}}(\mathbfz_s,s,\mathbf{m}))$. Specifically, for the step $s$ in the denosing process, the first Mamba block receives $[\mathbfz_s, \mathbf{m}]$ to calculate $y_{0,s}$ by Eq.~(\ref{eq:mdss_par}). Then the feature values of $y_{0,s}$ at the corresponding positions of the concatenated $\mathbf{m}$ are recovered to $\mathbf{m}$, which is fed to the following Mamba blocks to get $y_{0:\ell-1,s}$ iteratively.} The final denoised result $\mathbfz_{s-1}$ corresponds to the feature values of $y_{\ell-1,s}$ at the corresponding positions of $\mathbfz_s$. \myred{We further design a continuous-discrete-continuous (CDC) operation for explicit interaction on predicted hand waypoints in the reverse inference process, rather than being limited in the latent space that ignores the discrete nature of pixels in 2D image plane (see Sec.~\ref{sec:train_ref}).} \myblue{Ultimately, the denoised feature $\hat{\mathcal{F}}=f_{\scriptscriptstyle\text{Mamba}}(\mathbfz_1,1,\mathbf{m})=\{\hat{F}\}_{t=1}^{N_\text{f}}$ is fed to the trajectory decoder, which generates future hand trajectories in parallel.}

\begin{figure*}[t]
  \centering
  \includegraphics[width=0.81\linewidth]{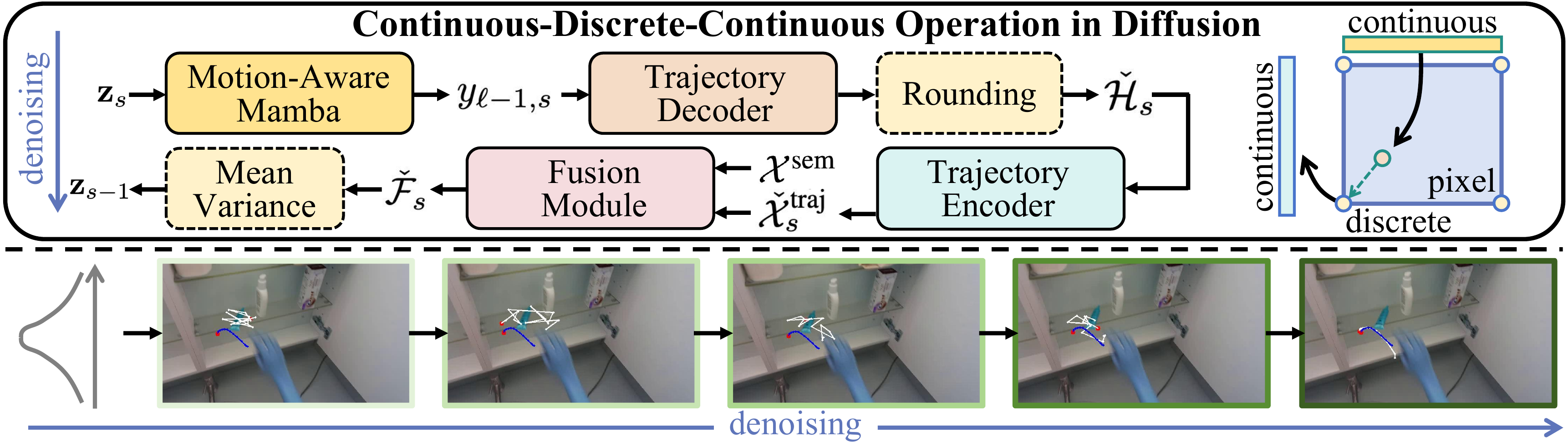}
  \vspace{-0.3cm}
  \caption{\myred{Continuous-discrete-continuous operation within each denoising step during the inference process (the upper part). We explicitly decode continuous intermediate latents to discrete hand waypoints (the lower part), and then encode them back to continuous latents for the following denoising.}}
  \label{fig:cdc}
  \vspace{-0.5cm}
\end{figure*}

\myblue{Note that we anchor $\mathbf{m}$ in Eq.~(\ref{eq:mdss_par}) for the inputs of all consecutive motion-aware Mamba blocks for two reasons:} 1) we respect the fact that egomotion is deterministic during the hand movement and should not be reconstructed as hand state features in the diffusion process (demonstrated in Sec.~\ref{sec:abla_on_mdss}), and 2) anchoring deterministic conditional information while denoising features enhances the stability of the optimization process~\cite{gong2022diffuseq,gong2023diffuseq,ma2024diff} and reduces the computation~\cite{chi2023diffusion}. In addition, following the previous works~\cite{gong2022diffuseq,ma2024diff}, we also anchor the past part of the latent features for each diffusion step to achieve conditional sequence modeling and apply both learnable positional embedding and temporal embedding before each denoising operation. \myred{Next, we will introduce MADiff's training strategy and how to integrate CDC operation (Fig.~\ref{fig:cdc}) into its inference process.}

\begin{figure}[t]
  \centering
  \includegraphics[width=0.9\linewidth]{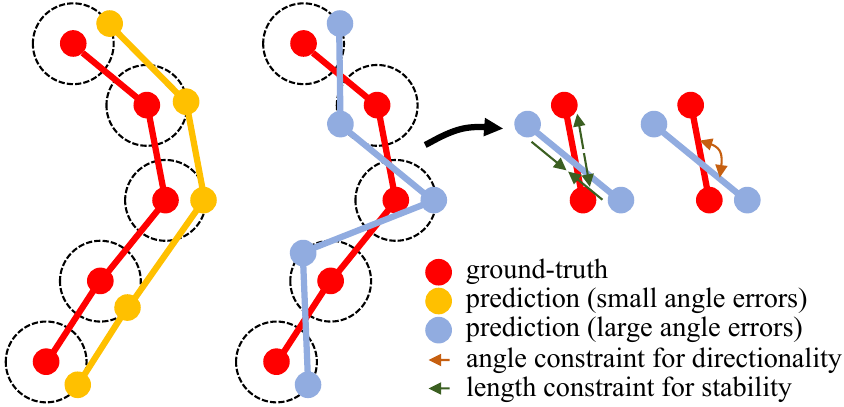}
  \caption{The motivation for designing new loss functions.}
  \label{fig:angle_loss}
  \vspace{-0.5cm}
\end{figure}

\subsection{MADiff Training and Inference}
\label{sec:train_ref}

\textbf{Training with New Losses:} We first use the same diffusion-related losses $\mathcal{L}_\text{VLB}$, trajectory displacement loss $\mathcal{L}_\text{dis}$, and regularization term $\mathcal{L}_\text{reg}$ as the previous work \cite{ma2024diff}:
\begin{align}
{\color{black}\mathcal{L}_\text{VLB}} &{\color{black}=} {\color{black}\sum_{s=2}^S \|\mathbf{z}_0 - f_{\scriptscriptstyle\text{Mamba}}(\mathbf{z}_s, s, \mathbf{m})\|^2 + \|\mathcal{F} - \hat{\mathcal{F}}\|^2,} \\
\mathcal{L}_\text{dis}&=\frac{1}{N_\text{f}}\sum_{t=1}^{N_\text{f}}D_\text{dis}(H_t,H_{t}^\text{gt}), \\
\mathcal{L}_\text{reg}&=\frac{1}{N_\text{f}}\sum_{t=1}^{N_\text{f}}D_\text{dis}(\tilde{H}_t,H_{t}^\text{gt}),
\label{eq:origin_loss}
\end{align}
where $D_\text{dis}(\cdot)$ represents the Euclidean distance between predicted hand waypoints and ground-truth ones, and $\tilde{H}_t$ denotes the output of the trajectory decoder with $\mathcal{F}$ as input. Moreover, we design two new loss functions, angle loss and length loss, to supervise our MADiff during the training process. As depicted in Fig.~\ref{fig:angle_loss}, the two predicted hand trajectories have the same displacement error, while the right case seems to be worse than the left one since it has ambiguous \textit{directionality} with large angle errors, and unreasonable \textit{stability} with large length errors. We argue that directionality and stability jointly reveal the causality and underlying human intention in the hand trajectory prediction task. Besides, they implicitly correspond to the potential physical model of hand motion and continuity constraints, closely associated with human habits. To promote the model capturing directionality and stability better, we propose the trajectory angle loss and length loss as follows:
\begin{align}
    \mathcal{L}_\text{angle}&=\frac{1}{N_\text{f}}\sum_{t=0}^{N_\text{f}-1}D_\text{cos}(H_{t+1}-H_t,H_{t+1}^\text{gt}-H_{t}^\text{gt}),\\
    \mathcal{L}_\text{len}&=\frac{1}{N_\text{f}}\sum_{t=0}^{N_\text{f}-1}D_\text{L2}(H_{t+1}-H_t,H_{t+1}^\text{gt}-H_{t}^\text{gt}),
\label{eq:angle_loss}
\end{align}
where $D_\text{cos}(\cdot)$ and $D_\text{L2}(\cdot)$ represent the cosine similarity and L2 norm of two input vectors respectively. \myblue{The total loss function to supervise MADiff is the weighted sum of all the above losses, depicted in Sec.~\myblue{A} of the supplementary material. The effectiveness of our new losses is experimentally demonstrated in Sec.~G of the supplementary material.}

\textbf{Inference with CDC Operation:} In the reference stage, we first sample noise $\mathcal{F}_\text{noise}=\{F_{t,\text{noise}}\}_{t=1}^{N_\text{f}}$ from a standard Gaussian distribution, and concatenate it with the past tokens $\mathcal{F}=\{F_t\}_{t=-N_\text{p}+1}^{0}$ along the time dimension to generate $\mathbfz_S$. Subsequently, the combination of motion-aware Mamba and our proposed CDC operation is adopted to predict future latent features by denoising $\mathbfz_S$ to $\mathbfz_0$. 
\myred{Specifically, prior to proceeding with the next denoising step $s-1$, the output of the stacked motion-aware Mamba blocks $y_{\ell-1,s}$, lying in the continuous latent space, is first converted to discrete hand waypoints $\check{\mathcal{H}}_{s}$ by the trajectory decoder.} We round the intermediate predictions $\check{\mathcal{H}}_{s}$ following the fact that the coordinates of hand waypoints on the 2D image grids are discrete. \myred{Since the denoising diffusion is implemented on the continuous latents, we subsequently project the discrete waypoints back to trajectory features $\check{\mathcal{X}}^{\text{traj}}_{s}$ by the trajectory encoder.} They are further fused with the vanilla semantic features $\mathcal{X}^{\text{sem}}$ by the fusion module in Fig.~\ref{fig:fusion_module} to derive $\check{\mathcal{F}}_{s}$, which is ultimately transformed to $\mathbfz_{s-1}$ for the following denoising steps. \myblue{The CDC operation's pipeline for diffusion-based HTP and intermediate discrete HTP results after rounding are shown in Fig.~\ref{fig:cdc}. 
As can be observed, the CDC operation constrains the optimization directions to the relationships between finite 2D discrete pixel coordinates. We expect this to reduce the difficulty of 2D HTP and accelerate the convergence of our denoising model. Moreover, if a predicted waypoint has already been correctly rounded to its ground-truth position, the loss gradient for this waypoint becomes negligible, allowing MADiff to concentrate on optimizing misaligned waypoints in challenging examples. Furthermore, analogous to the clamping trick in text generation~\cite{li2022diffusion,gong2022diffuseq}, the CDC operation restricts predicted waypoints to discrete image grids, preventing them from drifting far beyond image boundaries and thereby enhancing training stability.
}
\begin{figure}[t]
  \centering
  \includegraphics[width=0.97\linewidth]{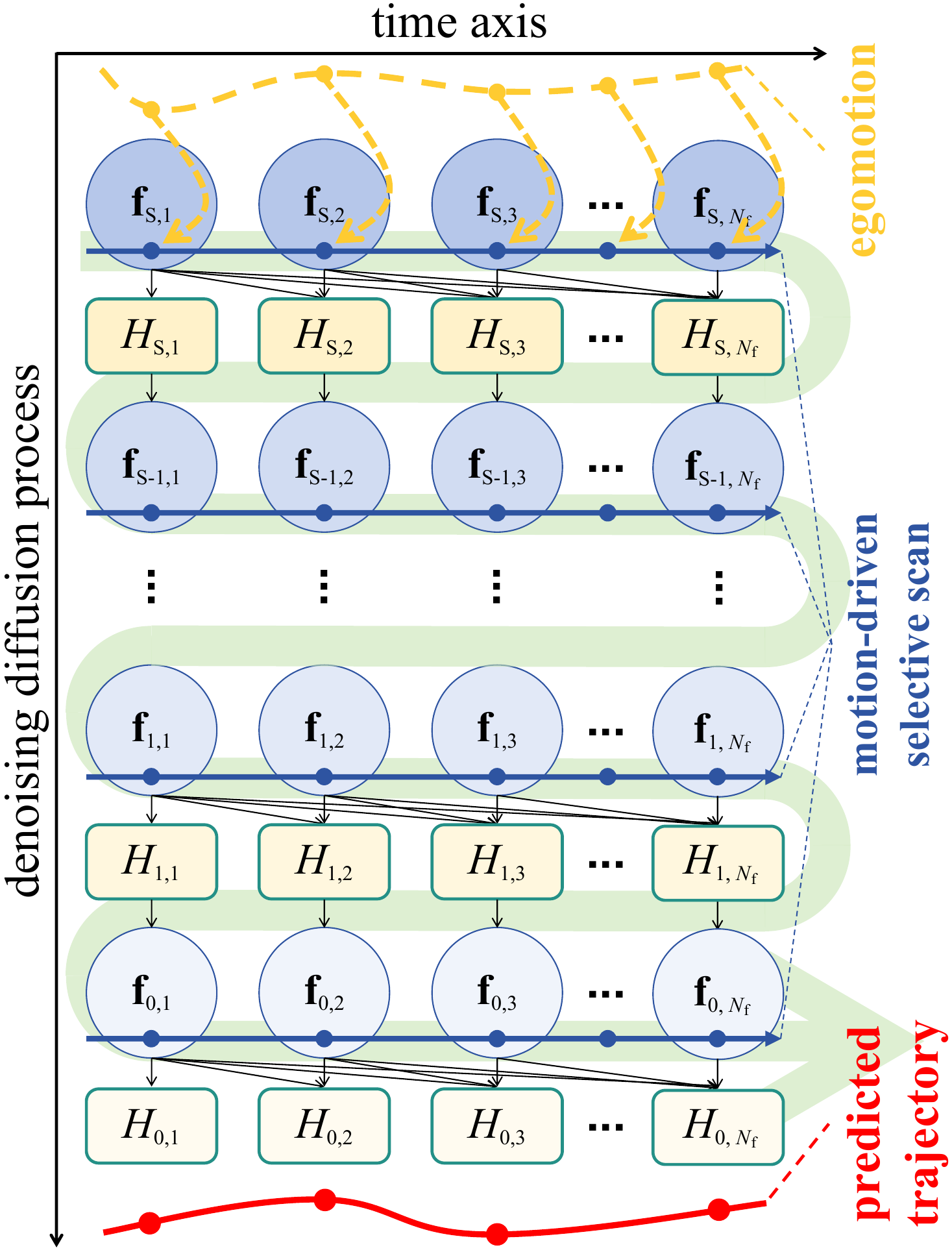}
  \vspace{-0.3cm}
  \caption{\myblue{MADiff iterates along both the denoising direction and the time axis to generate future hand trajectories, where MDSS is implemented following the temporal causality.}}
  \label{fig:mamba_denoising}
  \vspace{-0.5cm}
\end{figure}
Here we further show how MADiff bridges autoregressive (AR) models \cite{bao2023uncertainty,liu2022joint} and iterative non-autoregressive (iter-NAR) models \cite{ma2024diff}, which builds a novel generative paradigm for hand trajectory prediction. It captures the temporal causality along the time direction and maintains sufficient iteration in the denoising direction. We denote $\mathbf{f}_*$ as $\{\mathbf{f}_{S},\ldots,\mathbf{f}_0\}$ where $\mathbf{f}$ is the future part of $\mathbfz$, and $\mathcal{H}^\text{f}_*$ as $\{\mathcal{H}^\text{f}_S,\ldots,\mathcal{H}^\text{f}_1\}$ for brevity. \myblue{Considering egomotion guidance $\mathbf{m}$, the diffusion-based inference process of MADiff along with CDC operation can be formulated as:}

\begin{small}\color{black}
\begin{align}
&~p_{\scriptscriptstyle\text{MADiff}}(\mathcal{H}^\text{f}|\mathcal{H}^\text{p},\mathbf{m})  \nonumber \\
=& {\sum_{\mathcal{H}^\text{f}_*}\int_{\mathbf{f}_*}}{p(\mathcal{H}^\text{f}|\mathbf{f}_{0},\mathcal{H}^\text{p},\mathbf{m})}\prod_{s=S,\ldots,1}{p(\mathbf{f}_{s-1}|\mathcal{H}^\text{f}_s)p(\mathcal{H}^\text{f}_{s}|\mathbf{f}_s,\mathcal{H}^\text{p},\mathbf{m})} \nonumber \\ 
=& {\sum_{\mathcal{H}^\text{f}_*}\int_{\mathbf{f}_*}}p(\mathcal{H}^\text{f}_S|\mathbf{f}_S,\mathcal{H}^\text{p},\mathbf{m})\prod_{s=S-1,\ldots,0}{p(\mathcal{H}^\text{f}_s|\mathbf{f}_s,\mathcal{H}^\text{p},\mathbf{m})p(\mathbf{f}_s|\mathcal{H}^\text{f}_{s+1})} \nonumber \\ 
=&\sum_{\substack{\mathcal{H}^\text{f}_*}}p(\mathcal{H}^\text{f}_S|\mathbf{f}_S,\mathcal{H}^\text{p},\mathbf{m})\prod_{s=S-1,\ldots,0}{ {\int}_{\mathbf{f}_s}p(\mathcal{H}^\text{f}_s|\mathbf{f}_s,\mathcal{H}^\text{p},\mathbf{m})p(\mathbf{f}_s|\mathcal{H}^\text{f}_{s+1})}.
\label{eq:iter-nar1}
\end{align}
\end{small}
\begin{justify}
Then we marginalize over $\mathbf{f}$, and align the step $s$ with the general iteration number $k$ reversely, obtaining the iter-NAR form of MADiff:
\end{justify}
\begin{small}\color{black}
\begin{align}
&~p_{\scriptscriptstyle\text{MADiff}}(\mathcal{H}^\text{f}|\mathcal{H}^\text{p},\mathbf{m}) \nonumber \\ 
=&\sum_{\substack{\mathcal{H}^\text{f}_*}}p(\mathcal{H}^\text{f}_S|\mathbf{f}_S,\mathcal{H}^\text{p},\mathbf{m})\prod_{t=S-1,\ldots,0}{p(\mathcal{H}^\text{f}_s|\mathcal{H}^\text{f}_{s+1},\mathcal{H}^\text{p},\mathbf{m})} \nonumber \\ 
\equiv&\sum_{\mathcal{H}^\text{f}_1,\ldots,\mathcal{H}^\text{f}_{K-1}}{p(\mathcal{H}^\text{f}_1|\mathcal{H}^\text{p},\mathbf{m})\vphantom{\prod_{k=1,\ldots,K-1}}\prod_{k=1,\ldots,K-1}{p(\mathcal{H}^\text{f}_{k+1}|\mathcal{H}^\text{f}_k,\mathcal{H}^\text{p},\mathbf{m})}},
\label{eq:iter-nar2}
\end{align}
\end{small}
\begin{justify}
\vspace{-0.7cm}
\myblue{where $p(\mathcal{H}^\text{f}_1|\mathcal{H}^\text{p},\mathbf{m})$ and $p(\mathcal{H}^\text{f}_{k+1}|\mathcal{H}^\text{f}_k,\mathcal{H}^\text{p},\mathbf{m})$ correspond to the initial prediction and progressive full-context prediction of the general form of iter-NAR models respectively.}
Note that we predict hand waypoints $\mathcal{H}^\text{f}_{k}$ by the devised CDC operation in each step of the diffusion process rather than only denoised latents \cite{ma2024diff}, and thus Eq.~(\ref{eq:iter-nar2}) holds explicitly. Subsequently, we consider Mamba-based state transition of MADiff in Eq.~(\ref{eq:iter-nar2}), which can be an extension of the autoregressive scheme over $y$:
\end{justify}
\begin{small}\color{black}
\begin{align}
&~p_{\scriptscriptstyle\text{MADiff}}(\mathcal{H}^\text{f}|\mathcal{H}^\text{p},\mathbf{m}) \nonumber \\ 
\equiv&\sum_{\mathcal{H}^\text{f}_1,\ldots,\mathcal{H}^\text{f}_{K-1}}{p(\mathcal{H}^\text{f}_1|\mathcal{H}^\text{p},\mathbf{m})\vphantom{\prod_{k=1,\ldots,K-1}}\prod_{k=1,\ldots,K-1}{p(\mathcal{H}^\text{f}_{k+1}|\mathcal{H}^\text{f}_k,\mathcal{H}^\text{p},\mathbf{m})}} \nonumber \\ 
=&\sum_{\mathcal{H}^\text{f}_1,\ldots,\mathcal{H}^\text{f}_{K-1}}p(\mathcal{H}^\text{f}_1|\mathcal{H}^\text{p},\mathbf{m})\vphantom{\prod_{k=1,\ldots,K-1}}\prod_{k=1,\ldots,K-1}p(\mathcal{H}^\text{f}_{k+1}|y_k^{1:N_\text{p}+N_\text{f}})  \nonumber \\
&\quad\,p(y_k^1|\mathcal{H}^\text{f}_k,\mathcal{H}^\text{p},m_1)\prod_{i=1,\ldots,N_\text{p}+N_\text{f}-1}p(y_k^{i+1}|y_k^{1:i},\mathcal{H}^\text{f}_k,\mathcal{H}^\text{p},m_{i+1}),  
\label{eq:iter-nar3}
\end{align}
\end{small}
\begin{justify}
\vspace{-0.5cm}
where $i$ represents the time horizon where MDSS has been progressively implemented, and $p(y_k^1|\mathcal{H}^\text{f}_k,\mathcal{H}^\text{p},m_1)$ and $p(y_k^{i+1}|y_k^{1:i},\mathcal{H}^\text{f}_k,\mathcal{H}^\text{p},m_{i+1})$ represent the initial prediction and progressive left-context prediction of the general form of AR models respectively. Here we only consider one Mamba block with a single scan in Eq.~(\ref{eq:iter-nar3}) for brevity. $y_k^{i+1}$ is generated conditioned on both $\mathcal{H}^\text{f}_k$ and $\mathcal{H}^\text{p}$ because the projection functions in Eq.~(\ref{eq:proj_mat}) take the holistic latent sequence denoised by the previous steps as input, maintaining potential global-context constraints in the forward-only scan pattern. As the overall inference pipeline illustrated in Fig.~\ref{fig:mamba_denoising}, MADiff adopts the diffusion-based iter-NAR framework to keep sufficient iteration, and integrates motion-driven AR progress into each denoising step to capture temporal dependency orthogonal to the diffusion direction, which can serve as a foundation scheme for hand trajectory prediction and other time series forecasting tasks. 
\myblue{Since the future egomotion is unavailable during inference, we simply let $m_t (t>0)$ be $m_0$ for Eq.~(\ref{eq:mdss3}). We argue that reusing the last available egomotion feature for future time steps is a practically useful strategy, even in highly dynamic scenarios. Firstly, MDSS's unidirectional temporal modeling (from past to future) mitigates the effect of future egomotion features on the state transitions in the observed time steps. Therefore, our paradigm can still narrow the motion-related gaps introduced in Sec.~\ref{sec:intro} from past observations and improve trajectory prediction performance. Additionally, as the supervision signals from GT annotations remain consistent, MADiff is still optimized to predict reasonable 2D hand trajectories on the prediction canvas even without access to actual future egomotion. Besides, the last available egomotion feature can serve as a reasonable initial value for optimizing/correcting future egomotion features in future time step prediction. In Sec.~E of the
supplementary material, we also show that it is not trivial to explicitly predict reasonable future egomotion features, and inaccurate predictions introduce additional artifacts. Therefore, the choice of reusing the last available egomotion feature attends to the practical feasibility in our proposed paradigm.}

\end{justify}

\section{Experimental Results}
\label{sec:exp}

\subsection{Datasets}

We use five publicly available datasets to validate the superiority of our proposed MADiff, including Epic-Kitchens-55 (EK55) \cite{damen2018scaling}, Epic-Kitchens-100 (EK100) \cite{damen2022rescaling}, EGTEA Gaze+ (EG) \cite{li2018eye}, EgoPAT3D-DT \cite{li2022egocentric,bao2023uncertainty}, and H2O-PT \cite{kwon2021h2o,bao2023uncertainty}. We use the EK55 and EK100 datasets following the setups of OCT \cite{liu2022joint} and Diff-IP2D \cite{ma2024diff}, where we sample past $N_\text{p}= 10$ frames (2.5\,s) to forecast hand waypoints in future $N_\text{f}= 4$ frames (1.0\,s), both at 4 FPS. For the EG dataset, $N_\text{p}= 9$ frames (1.5\,s) are used for $N_\text{f}= 3$ hand trajectory predictions (0.5\,s) at 6 FPS. Following the setups of USST \cite{bao2023uncertainty}, we use the fixed ratio 60\% by default to split the past and future sequences for both EgoPAT3D-DT and H2O-PT at 30 FPS. Sec.~\myblue{C} in the supplementary material further presents the effects of different observation ratios in the two datasets. EgoPAT3D-DT contains both seen and unseen scenes, where the unseen scenes are only used for testing. The numbers of video clips in the training, validation, and testing splits for different datasets used in the following experiments are shown in Tab.~\ref{tab:splits}. According to the specific annotations in different datasets, we use the image plane of the last observation as the prediction canvas on EK55, EK100, and EG, and instead use the image plane of the first observation as the canvas on EgoPAT3D-DT and H2O-PT.

\begin{table}[t]
\small
\centering
\setlength{\tabcolsep}{1.4mm}
\caption{Dataset splits for hand trajectory prediction. EgoPAT3D-DT has both seen/unseen scenarios for testing.}
\vspace{-0.2cm}
\begin{tabular}{lccccc}
\toprule
Dataset & \begin{tabular}[c]{@{}c@{}}EK55\end{tabular} & \begin{tabular}[c]{@{}c@{}}EK100\end{tabular} & EG & \begin{tabular}[c]{@{}c@{}}EgoPAT3D-DT \end{tabular}  & H2O-PT \\ \midrule
training   & 8523       & 24148       & 1880       & 6356       & 8203      \\ 
validation   & 241       & 401       & 69        & 846       & 1735       \\ 
testing   & 1894       & 3513       & 442       & 1605/2334       & 3715       \\ \bottomrule
\end{tabular}
\label{tab:splits}
\vspace{-0.5cm}
\end{table}

\subsection{MADiff Configurations}
\label{sec:madiff_config}

We use GLIP \cite{Li_2022_CVPR} as the foundation model to generate the semantic feature with a size of $256\times 7\times 12$ for each frame, which is then transformed to a feature vector with a size of $512$ in the fusion module. In this work, we use the GLIP version with a Swin-Large backbone \cite{liu2021swin} as well as BERT (base-uncased) \cite{devlin2018bert} to encode the text prompt. The trajectory encoder embeds each 2D hand waypoint to a feature vector with a size of $512$. The output token of the fusion module for each timestamp is a feature vector with a size of $512$. The homography encoder converts each $3\times 3$ homography matrix to a feature vector with a size of $512$. Although MADiff uses SIFT+RANSAC to calculate the homography matrix by default, we provide an additional study on its robustness to multiple homography estimation algorithms in Sec.~\myblue{D} of the
supplementary material. As to the diffusion process, the total number of steps is set to 1000. The square-root noise schedule in Diffusion-LM \cite{li2022diffusion} is adopted here for the forward diffusion process. We use 6 stacked motion-aware Mamba blocks with convolutional kernel size $d\_conv=2$, hidden state expansion $expand=1$, and hidden dimension $d\_state=16$ as the denoising model. 
The numbers of diffusion steps and Mamba blocks are both selected according to the ablation study in Sec.~\ref{sec:abla_mamba_diff}.
We train MADiff using AdamW optimizer \cite{kingma2014adam} with a learning rate of 2e-4 for 20 epochs on Epic-Kitchens, and with a learning rate of 1e-4 for 400 epochs on both EgoPAT3D-DT and H2O-PT. Training and inference are both operated on 2 A100 GPUs.

\subsection{Baseline Selection}

For the EK55, EK100, and EG datasets, we follow the previous work \cite{ma2024diff} and choose CVH \cite{ma2024diff}, Seq2Seq \cite{sutskever2014sequence}, FHOI \cite{liu2020forecasting}, OCT \cite{liu2022joint}, USST \cite{bao2023uncertainty}, and Diff-IP2D \cite{ma2024diff} as the baselines. For the EgoPAT3D-DT and H2O-PT datasets, we select the baselines including CVH \cite{ma2024diff}, DKF \cite{krishnan2015deep}, RVAE \cite{leglaive2020recurrent}, DSAE \cite{li2018disentangled}, STORN \cite{bayer2014learning}, VRNN \cite{chung2015recurrent}, SRNN \cite{fraccaro2016sequential}, EgoPAT3D \cite{li2022egocentric}, AGF \cite{yuan2021agentformer}, OCT \cite{liu2022joint}, ProTran \cite{tang2021probabilistic}, USST \cite{bao2023uncertainty}, and Diff-IP2D \cite{ma2024diff}, where we partially refer to the baselines of the previous work \cite{bao2023uncertainty}. Note that we use the 2D version of USST since there is no available 3D information for the prediction task in this work. We borrow partial quantitative results for these baselines from the previous works \cite{ma2024diff,bao2023uncertainty} since we keep the same experimental configurations as them. 

\begin{figure}[t]
  \centering
  \includegraphics[width=1\linewidth]{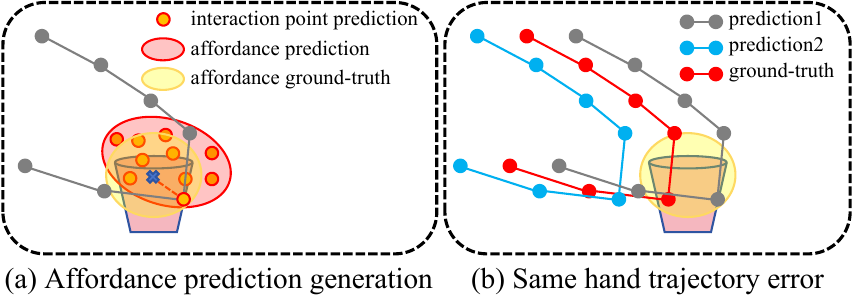}
  \vspace{-0.5cm}
  \caption{We evaluate the distribution of ``interaction points'' of predicted hand trajectories, revealing the interaction relationship between hands and active objects.}
  \label{fig:new_metric}
  \vspace{-0.5cm}
\end{figure}

\begin{table*}[t]
\small
\setlength{\tabcolsep}{15pt}
\center
\caption{Comparison of performance on hand trajectory prediction on the EK55, EK100, and EG datasets. Best and secondary results are viewed in \textbf{bold black} and \mygreen{bold green} colors respectively.}
\vspace{-0.2cm}
\begin{tabular}{l|ccc|ccc|ccc}
\toprule
\multicolumn{1}{l|}{\multirow{2}{*}{Approach}}   & \multicolumn{3}{c|}{EK55} & \multicolumn{3}{c|}{EK100}  & \multicolumn{3}{c}{EG} \\ \cmidrule{2-10} 
\multicolumn{1}{c|}{}                                                                               & WDE\,$\downarrow$  &  & FDE\,$\downarrow$ & WDE\,$\downarrow$  &  & FDE\,$\downarrow$ & WDE\,$\downarrow$  &  & FDE\,$\downarrow$   \\ \cmidrule{1-10}                                                                   CVH \cite{ma2024diff} & 0.636   &   & 0.315     &  0.658  &   & 0.329    & 0.689  &   & 0.343    \\
Seq2Seq \cite{sutskever2014sequence}  & 0.505  &    & 0.212      &  0.556   &  & 0.219    & 0.649 &    & 0.263   \\ 
FHOI \cite{liu2020forecasting}   & 0.589    &     & 0.307     & 0.550   &     & 0.274    & 0.557   &  & 0.268   \\ 
OCT \cite{liu2022joint} & {0.446}   &    & {0.208}     & {0.467}   &    & {0.206}    & {0.514}    &   & {0.249}   \\ 
USST \cite{bao2023uncertainty}  & 0.458   &   & 0.210           & 0.475   &    & {0.206}  & 0.552 &   & 0.256   \\ 
Diff-IP2D \cite{ma2024diff}  & \mygreen{0.411}   &   & \mygreen{0.181}           & \mygreen{0.407}   &    & \mygreen{0.187}  & \mygreen{0.478} &   & \mygreen{0.211}   \\ 
\rowcolor{lightgray}
MADiff (ours)  & \textbf{0.374}   &    & \textbf{0.169}     &  \textbf{0.387}  &  & \textbf{0.176}    &  \textbf{0.454}  &  & \textbf{0.203}          \\ \bottomrule
\end{tabular}
\label{tab:compare_hand_epic_eg}
\vspace{-0.1cm}
\end{table*}

\begin{table*}[t]
\small
\setlength{\tabcolsep}{10.9pt}
\center
\caption{Comparison between MADiff and the other baselines supervised by affordance labels with our new metrics on the EK55, EK100, and EG datasets. Best and secondary results are viewed in \textbf{bold black} and \mygreen{bold green} respectively.}
\vspace{-0.2cm}
\begin{tabular}{l|ccc|ccc|ccc}
\toprule
\multicolumn{1}{l|}{\multirow{2}{*}{Approach}}   & \multicolumn{3}{c|}{EK55} & \multicolumn{3}{c|}{EK100}  & \multicolumn{3}{c}{EG} \\ \cmidrule{2-10} 
\multicolumn{1}{c|}{}                                                                               & SIM\,$\uparrow$  & AUC-J\,$\uparrow$  & NSS\,$\uparrow$ & SIM\,$\uparrow$  & AUC-J\,$\uparrow$  & NSS\,$\uparrow$ & SIM\,$\uparrow$  & AUC-J\,$\uparrow$  & NSS\,$\uparrow$   \\ \cmidrule{1-10}                                                         
FHOI$^{\dag}$\cite{liu2020forecasting}   & 0.127 & 0.503 & 0.455 & 
 0.110 & 0.529 & 0.386 & 0.102 & 0.497 & 0.352  \\ 
OCT$^{\dag}$\cite{liu2022joint}  & 0.190 & 0.657 & 0.750 & 0.167  &  0.642 &  0.578 & 0.181 & 0.614 & 0.642    \\ 
Diff-IP2D$^{\dag}$\cite{ma2024diff}  & {0.195} & {0.663} & {0.764} & {0.185} & {0.660} & {0.796} & {0.208} & {0.651} & {0.694}  \\  \midrule
FHOI \cite{liu2020forecasting}   & 0.156 & 0.612 & 0.574 & 
 0.139 & 0.560 & 0.449 & 0.144 & 0.569 & 0.427  \\ 
OCT \cite{liu2022joint}  & 0.205 & 0.660 & 0.802 & 0.197  & \mygreen{0.672} &  0.710 & 0.204 & 0.670 & 0.810    \\ 
Diff-IP2D \cite{ma2024diff}  & \mygreen{0.210} & \mygreen{0.665} & \mygreen{0.856} & \mygreen{0.221} & {0.667} & \mygreen{0.931} & \mygreen{0.235} & \mygreen{0.677} & \mygreen{0.845}  \\ 
\rowcolor{lightgray}
MADiff (ours)  & \textbf{0.241}   & \textbf{0.670}    & \textbf{1.03}     &  \textbf{0.233}  & \textbf{0.680}  & \textbf{1.02}    &  \textbf{0.240}  & \textbf{0.704} & \textbf{0.992}          \\ \bottomrule
\end{tabular}
\label{tab:compare_affordance}
\\
\begin{flushleft}
\scriptsize
$^{\dag}$\,We use the baselines' predicted affordance instead of ground-truth ones to calculate our new metrics since they are explicitly supervised by object affordance labels.
\end{flushleft}
\vspace{-0.4cm}
\end{table*}

\begin{table*}[t]
\small
\setlength{\tabcolsep}{14.8pt}
\center
\caption{Comparison of performance on hand trajectory prediction on the EgoPAT3D-DT and H2O-PT datasets. Best and secondary results are viewed in \textbf{bold black} and \mygreen{bold green} colors respectively.}
\vspace{-0.2cm}
\begin{tabular}{l|ccc|ccc|ccc}
\toprule
\multicolumn{1}{l|}{\multirow{2}{*}{Approach}}   & \multicolumn{3}{c|}{EgoPAT3D-DT (seen)} & \multicolumn{3}{c|}{EgoPAT3D-DT (unseen)}  & \multicolumn{3}{c}{H2O-PT} \\ \cmidrule{2-10} 
\multicolumn{1}{c|}{}                                                                               & ADE\,$\downarrow$  &  & FDE\,$\downarrow$ & ADE\,$\downarrow$  &  & FDE\,$\downarrow$ & ADE\,$\downarrow$  &  & FDE\,$\downarrow$   \\ \cmidrule{1-10}                 
CVH \cite{ma2024diff} & 0.180 &   & 0.230   & 0.188     &    & 0.221   & 0.206    &   & 0.208        \\
DKF \cite{krishnan2015deep} & 0.157 &   & 0.150   & 0.133     &    & 0.239   & 0.211    &   & 0.185        \\
RVAE$^{*}$ \cite{leglaive2020recurrent}  & 0.121 &   & 0.152   & 0.109       &    & 0.201   & 0.103     &  & 0.127       \\ 
DSAE \cite{li2018disentangled}   & 0.143  &  & 0.144    &  0.131     &    & 0.233    & 0.059    &    & 0.076     \\ 
STORN \cite{bayer2014learning} & 0.083  &  & 0.145    & 0.070      &    & 0.266   & 0.053     &     & 0.076      \\ 
VRNN \cite{chung2015recurrent}  & 0.083  &  & 0.155   & 0.070      &    & 0.237    & 0.050   &   & \textbf{0.068}    \\ 
SRNN$^{*}$ \cite{fraccaro2016sequential}  & \mygreen{0.079} &  & 0.157   & 0.067     &    &  0.198   & 0.062   &  & 0.107    \\ 
EgoPAT3D* \cite{li2022egocentric}  & \mygreen{0.079}  &  & --   & 0.068        &    & --   & 0.050   &  & 0.084      \\ 
AGF \cite{yuan2021agentformer}  & 0.099  &  & --   & 0.087       &    & --    & 0.081  &  & 0.146    \\ 
OCT$^{*}$ \cite{liu2022joint}  & 0.108  &   & {0.122}   &  0.091          &    &  0.147   & 0.387   &  & 0.381    \\ 
ProTran \cite{tang2021probabilistic}  & 0.135   &   & 0.134   & 0.107           &    & \textbf{0.049}    & 0.109  &  & 0.123     \\ 
USST \cite{bao2023uncertainty}  & {0.082}  &  & \mygreen{0.118}   & \mygreen{0.060}           &    & {0.087}    & \mygreen{0.040}  &     & \textbf{0.068}     \\ 
Diff-IP2D \cite{ma2024diff}   & {0.080}  &  & 0.130   & {0.066}           &    & {0.087}    & {0.042}  &     & \mygreen{0.074}     \\ 
\rowcolor{lightgray}
MADiff (ours)  & \textbf{0.065}   &    & \textbf{0.105}     &  \textbf{0.054}  &  & \mygreen{0.086}    &  \textbf{0.039}  &  & \textbf{0.068}          \\ \bottomrule
\end{tabular}
\label{tab:compare_hand_egopat_h2o}
\\
\vspace{-0.16cm}
\begin{flushleft}
\scriptsize
$^*$\,The baselines are re-evaluated according to the erratum: \url{https://github.com/oppo-us-research/USST/commit/beebdb963a702b08de3a4cf8d1ac9924b544abc4}.
\end{flushleft}
\vspace{-0.7cm}
\end{table*}

\subsection{Evaluation on Hand Trajectory Prediction}
\label{sec:eval_htp}

We evaluate the weighted displacement error (WDE) and the final displacement error (FDE) of our MADiff and all the baselines on the EK55, EK100, and EG datasets following Diff-IP2D \cite{ma2024diff}, and post the averaged displacement error (ADE) and FDE on the EgoPAT3D-DT and H2O-PT datasets following USST \cite{bao2023uncertainty}. 
\myblue{Compared to ADE, WDE introduces additional time-dependent weights to the errors at different future time steps, which is calculated by:
\begin{small}
\begin{align}
    \text{WDE}=\frac{1}{N_\text{f}}\sum_{t=1}^{N_\text{f}}\frac{t}{N_\text{f}}D_\text{L2}(H_t,H_t^\text{gt}). 
\label{eq:wde}
\end{align}
\end{small}
\hspace{-0.27cm} Thus, the later waypoints have relatively larger weights, following the fact that the general knowledge of ``post-contact trajectories'' extracted from human videos is particularly important for downstream tasks~\cite{ma2024diff}.}
Moreover, we further design a new metric to better evaluate the interaction between the hand and the next active objects, which is showcased in Fig.~\ref{fig:new_metric}(a). For each video clip, we generate 10 possible hand trajectory predictions $\{\mathcal{H}^\text{f}\}_{n=1}^{10}$, and select the waypoint closest to the affordance center $O^\text{f}$ of the next active object as the ``interaction point'' for each trajectory by
\begin{align}
    H_n^{\text{ip}}=\text{min}_t D_\text{dis}(\mathcal{H}_n, O^\text{f}).
\end{align}
Then we calculate the mixture of Gaussians of the 10 interaction points $\{H_n^{\text{ip}}\}_{n=1}^{10}$ as affordance prediction. The similarity between affordance prediction and affordance ground-truth is ultimately evaluated by Similarity Metric (SIM) \cite{swain1991color}, AUC-Judd (AUC-J) \cite{judd2009learning}, and Normalized Scanpath Saliency (NSS) \cite{peters2005components}. Our proposed new metric can distinguish the quality of predictions with similar displacement errors shown in Fig.~\ref{fig:new_metric}(b) based on the fact that the future hand movement always changes the state of an object by using or manipulating it \cite{grauman2022ego4d}. Note that affordance similarity of predicted hand trajectories can only be evaluated on the datasets EK55, EK100, and EG which provide ground-truth affordance labels from annotated contact points~\cite{liu2022joint}.

\begin{figure}[t]
  \centering
  \includegraphics[width=1\linewidth]{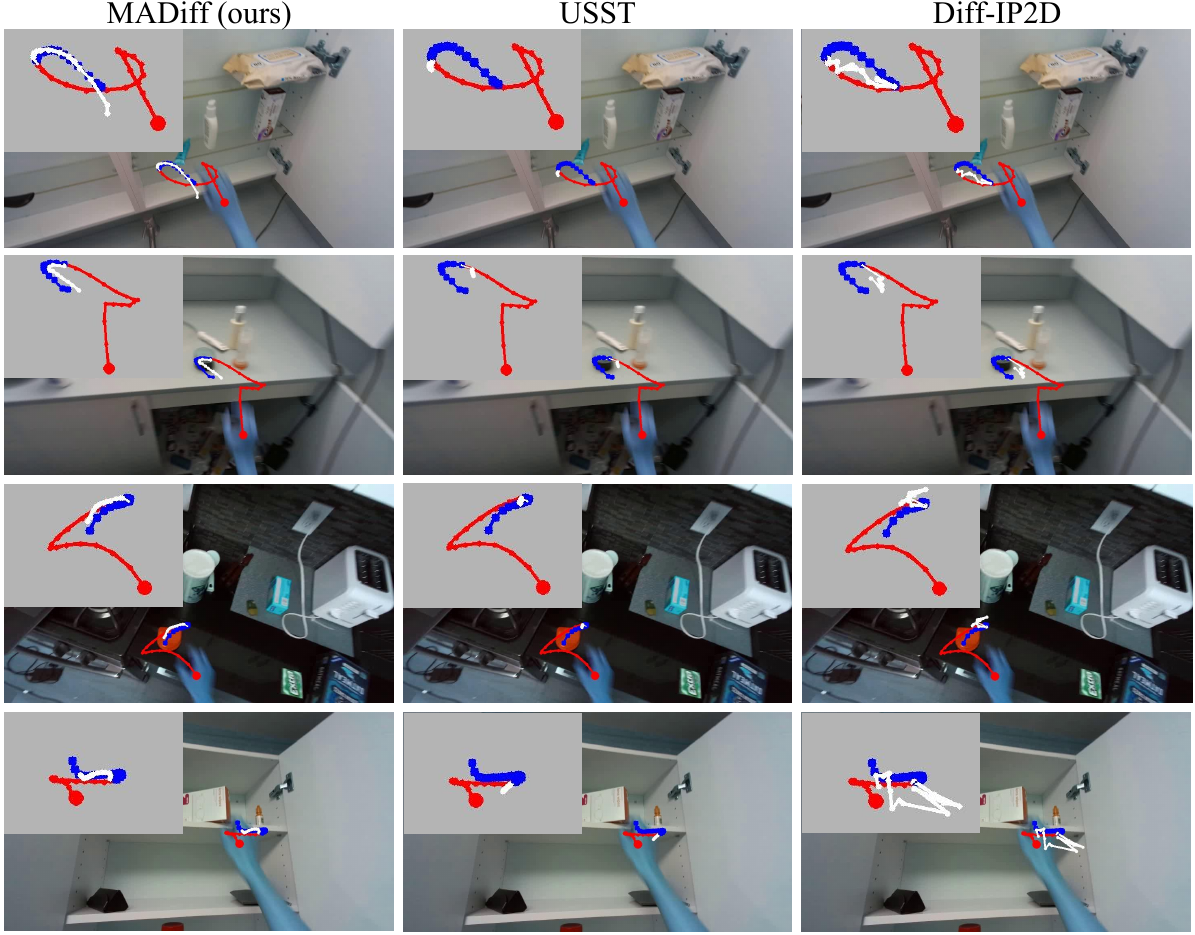}
  \caption{\myblue{Visualization of predicted hand trajectories in the selected hard cases. The future hand waypoints from ground-truth labels and HTP approaches are represented in blue and white, respectively, with the first frame of each sequence as the canvas. The past hand waypoints are represented in red, and the first history waypoint is enlarged to highlight the alignment between trajectories and hands. We reverse RGB values of each image to display the arm's positions more clearly (akin to a blue mask on the moving arm).}}
  \label{fig:partial_prediction}
  \vspace{-0.3cm}
\end{figure}

\begin{figure}[t]
  \centering
  \includegraphics[width=0.99\linewidth]{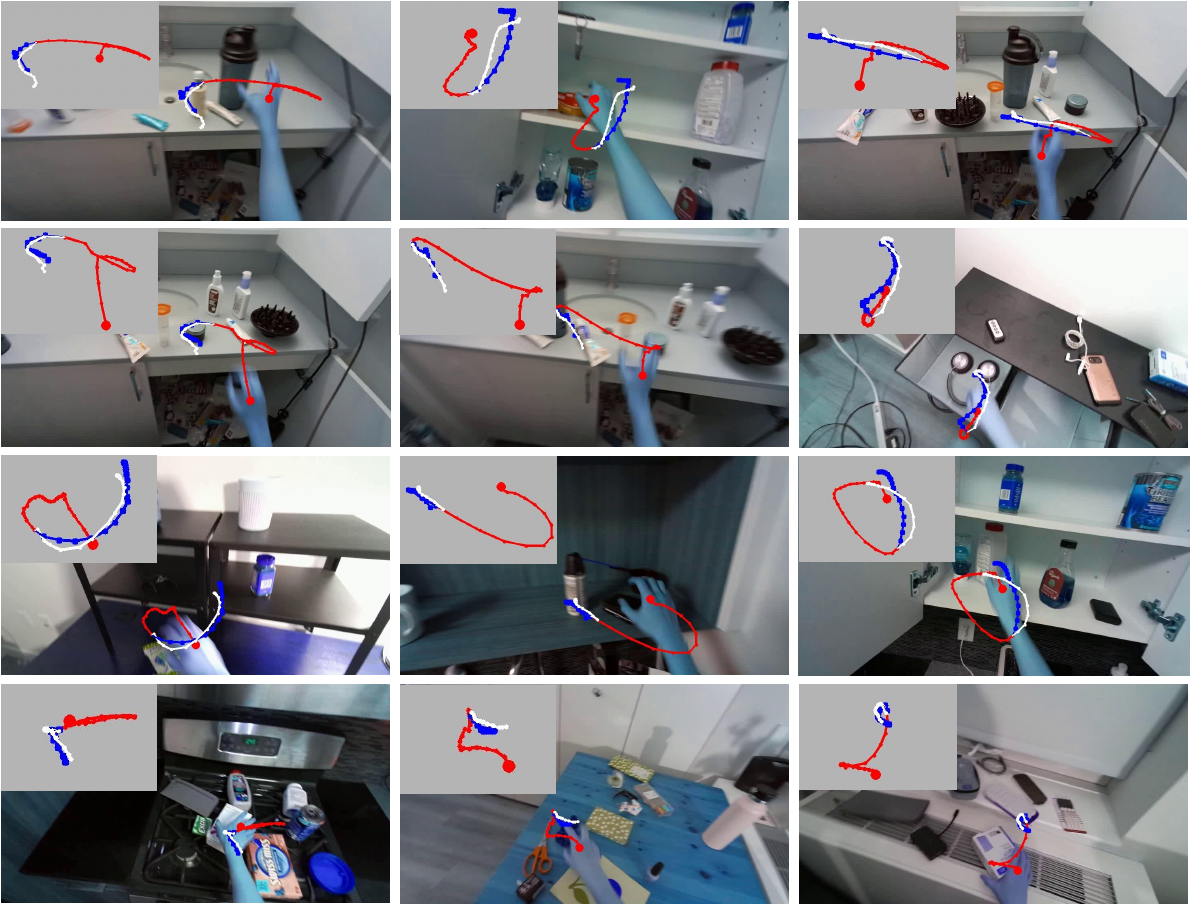}
  \caption{\myred{Additional illustrations of HTP by our MADiff.}}
  \label{fig:more_ill}
  \vspace{-0.6cm}
\end{figure}

We present the comparison results on the EK55, EK100, and EG datasets in Tab.~\ref{tab:compare_hand_epic_eg} and Tab.~\ref{tab:compare_affordance}. Tab.~\ref{tab:compare_hand_egopat_h2o} shows the comparison results on the EgoPAT3D-DT and H2O-PT datasets. Note that we implement zero-shot transfer from Epic-Kitchens to the EG dataset, from EgoPAT3D-DT (seen) to EgoPAT3D-DT (unseen), to validate the generalization ability on diverse scenes. As can be seen, MADiff outperforms all the baselines on the EK55, EK100, and EG datasets, and generates comparable (top 2) predictions on EgoPAT3D-DT and H2O-PT, which suggests MADiff's good hand trajectory prediction performance. The comparison results on EG and EgoPAT3D-DT (unseen) also demonstrate MADiff's strong generalization ability while facing new human activity environments. As to the evaluation on our new metrics in Tab.~\ref{tab:compare_affordance}, MADiff without affordance supervision still generates the most reasonable interaction distribution against other baselines supervised by object affordance annotations. This indicates that MADiff is capable of capturing potential relationships between hands and active objects. We provide the visualization of predicted hand trajectories from state-of-the-art baselines and MADiff on EgoPAT3D-DT in Fig.~\ref{fig:partial_prediction}. More illustrations of MADiff predictions can be found in Fig.~\ref{fig:more_ill}, Fig.~\myblue{A} and Fig.~\myblue{B} of the supplementary material.

\begin{table*}[t]
\small
\setlength{\tabcolsep}{3.6pt}
\center
\caption{\myblue{Ablation study on motion-driven selective scan, where the scores for the best performance are bolded.}}
\vspace{-0.2cm}
\begin{tabular}{l|ccc|ccc|ccc}
\toprule
\multicolumn{1}{l|}{\multirow{2}{*}{Approach}}   & \multicolumn{3}{c|}{EgoPAT3D-DT (seen)} & \multicolumn{3}{c|}{EgoPAT3D-DT (unseen)}  & \multicolumn{3}{c}{H2O-PT} \\ \cmidrule{2-10} 
\multicolumn{1}{c|}{}                                                                               & ADE\,$\downarrow$  &  & FDE\,$\downarrow$ & ADE\,$\downarrow$  &  & FDE\,$\downarrow$ & ADE\,$\downarrow$  &  & FDE\,$\downarrow$   \\ \cmidrule{1-10}  
\myblue{oracle (using GT future egomotion features)}   & \myblue{\textit{0.064}}  &  & \myblue{\textit{0.102}}  & \myblue{\textit{0.054}}    &    & \myblue{\textit{0.084}}           & \myblue{\textit{0.037}}  &   & \myblue{\textit{0.064}}   \\ \midrule
\myblue{v1 (totally removing egomotion features in MDSS)}   & 0.067  &  & 0.113  & 0.059    &    & 0.098           & 0.042  &   & 0.080   \\ 
\myblue{v2 (adding egomotion features to fusion module output)}   & 0.119  &   & 0.156       & 0.102   &    &  0.135  & 0.046  &   & 0.086   \\ 
\myblue{v3 (replacing concatenation with summation in MDSS)}   & 0.069  &   & 0.110       & 0.056   &    &  0.089  & 0.044  &   & 0.080   \\ 
\myblue{v4 (replacing unidirectional MDSS with bidirectional scan)}  & 0.070  &   & 0.109       & 0.057    &    & 0.089   & 0.042  &   & 0.072    \\ 
\rowcolor{lightgray}
MADiff (ours)  & \textbf{0.065}   &    & \textbf{0.105}     &  \textbf{0.054}  &  & \textbf{0.086}    &  \textbf{0.039}  &  & \textbf{0.068}          \\ \bottomrule
\end{tabular}
\label{tab:ala_on_mdss}
\vspace{-0.5cm}
\end{table*}

\begin{figure}[t]
  \centering
  \includegraphics[width=0.9\linewidth]{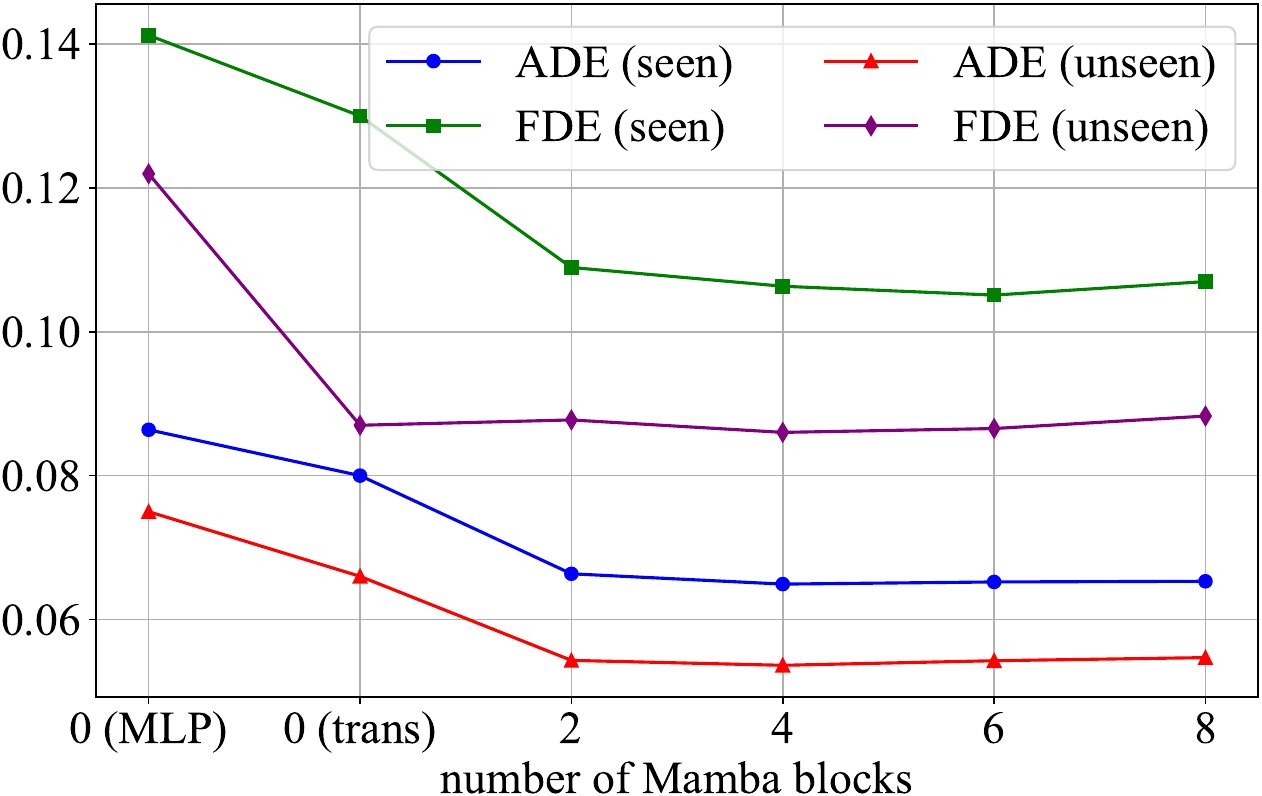}
  \vspace{-0.2cm}
  \caption{Trajectory displacement errors vs. numbers of our devised motion-aware Mamba blocks.}
  \label{fig:ablation_on_mamba_blocks}
  \vspace{-0.5cm}
\end{figure}

\subsection{Ablation Study on Motion-Driven Selective Scan}
\label{sec:abla_on_mdss}

This experiment is conducted on the EgoPAT3D-DT and H2O-PT datasets to show the effectiveness of our proposed motion-driven selective scan. 
\myblue{
We directly remove motion guidance to build the baseline MADiff agnostic to egomotion (version 1), where the concatenation between the egomotion features with latents in Eq.~(\ref{eq:mdss_par}) is omitted, reverting MDSS to vanilla selective scan. Building on version 1, we merge egomotion features with the fusion module's output via MLPs (version 2), replacing MDSS-based guidance with early feature merging before the denoising diffusion process. We also provide a baseline that replaces concatenation in Eq.~(\ref{eq:mdss_par}) with summation (version 3). To further assess the role of unidirectional temporal causality, we introduce a bidirectional Mamba~\cite{wang2024mamba} baseline (version 4), which scans sequential latents in two opposite directions. Please refer to the illustration in Sec.~I of the supplementary material for more intuitive construction processes of these baselines. Ultimately, we conduct an oracle baseline that integrates the GT future egomotion features into our MDSS. It provides valuable reference performance by introducing leaked future human motion patterns. The experimental results are shown in Tab.~5. The vanilla MDSS in our MADiff results in performance closest to the counterpart determined by the oracle baseline with GT future egomotion.
}
When comparing version 1 with our vanilla MADiff, it can be seen that motion guidance helps to reduce ADE and FDE on both datasets, which indicates that our proposed motion-driven selective scan narrows the motion-related gaps and concurrently considers the entangled hand motion and egomotion patterns. The enhancement from MDSS is more significant on FDE than ADE, which corresponds to the fact that there is an accumulated motion gap between a later observation and the canvas observation (i.e., the first observation for EgoPAT3D-DT and H2O-PT). Version 2 has the worst prediction performance among all the baselines, revealing that egomotion can only be used as auxiliary information within the diffusion process rather than brutally being fused with semantic and trajectory features that need to be optimally reconstructed by the denoising model, which has been claimed in Sec.~\ref{sec:mamba_diff}.  

In addition, MADiff with concatenation for motion guidance outperforms version 3 with summation operation. This suggests that the feature update from egomotion homography should not be directly added to the original state transition process without reweighting by the input-dependent projection parameters in Eq.~(\ref{eq:mdss1}). Version 4 has worse HTP performance than vanilla MADiff even though it applies bidirectional Mamba. The reason could be that traversing the latent sequence in the opposite direction with MDSS is analogous to strictly reversing the causal relationship and human motion patterns, leading to unreasonable denoising during training and inference. Thus, we advocate a forward-only scan with global-context constraints (Eq.~(\ref{eq:iter-nar3})) in our motion-aware Mamba rather than the bidirectional one.

\begin{figure}[t]
  \centering
  \includegraphics[width=1\linewidth]{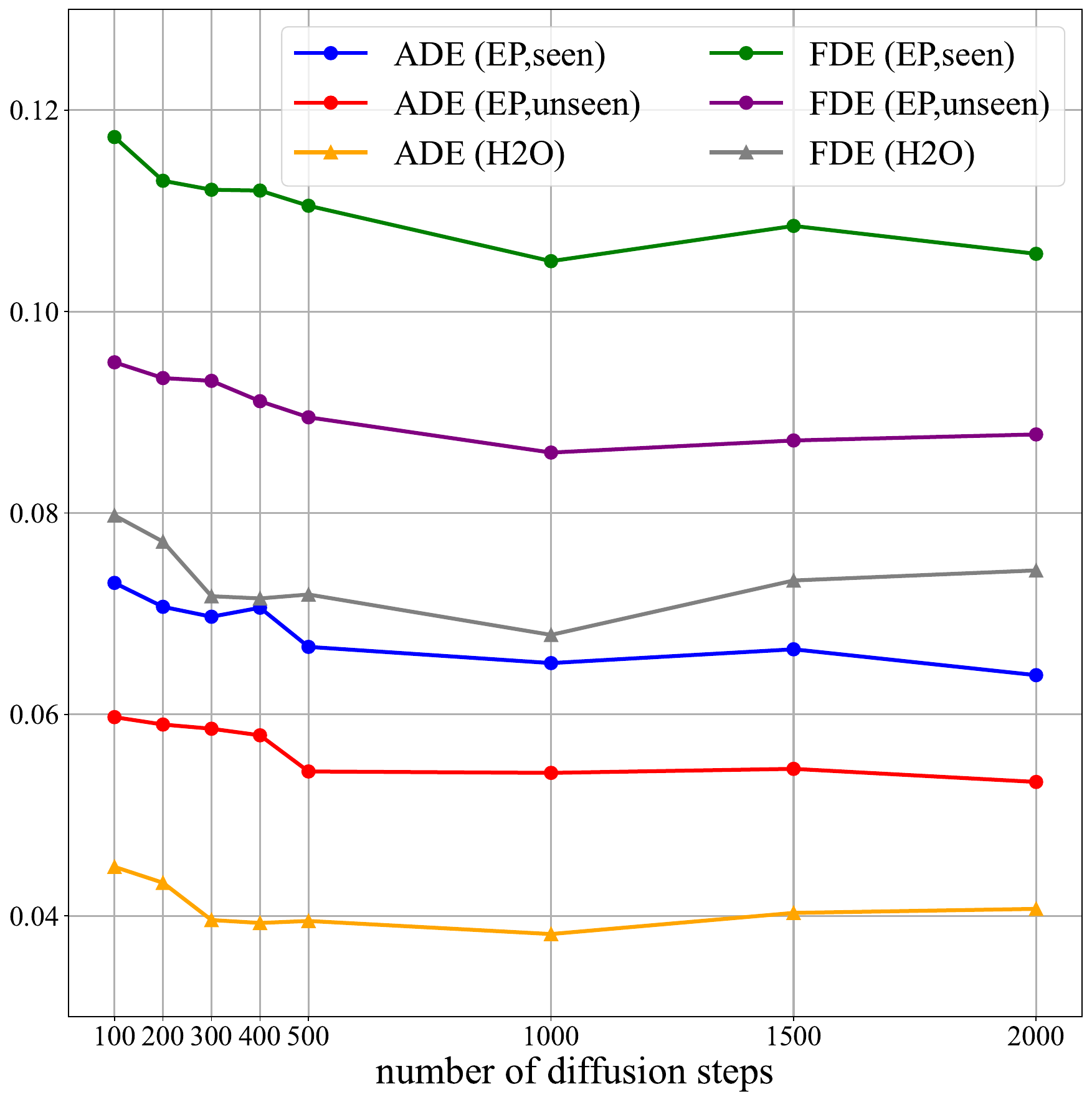}
  \vspace{-0.3cm}
  \caption{\myblue{Trajectory displacement errors vs. numbers of diffusion steps in MADiff.}}
  \label{fig:ablation_on_diffusion_steps}
  \vspace{-0.7cm}
\end{figure}

\begin{figure}[t]
    \centering
    \begin{subfigure}[t]{1\linewidth}
        \centering
        \includegraphics[width=0.9\linewidth]{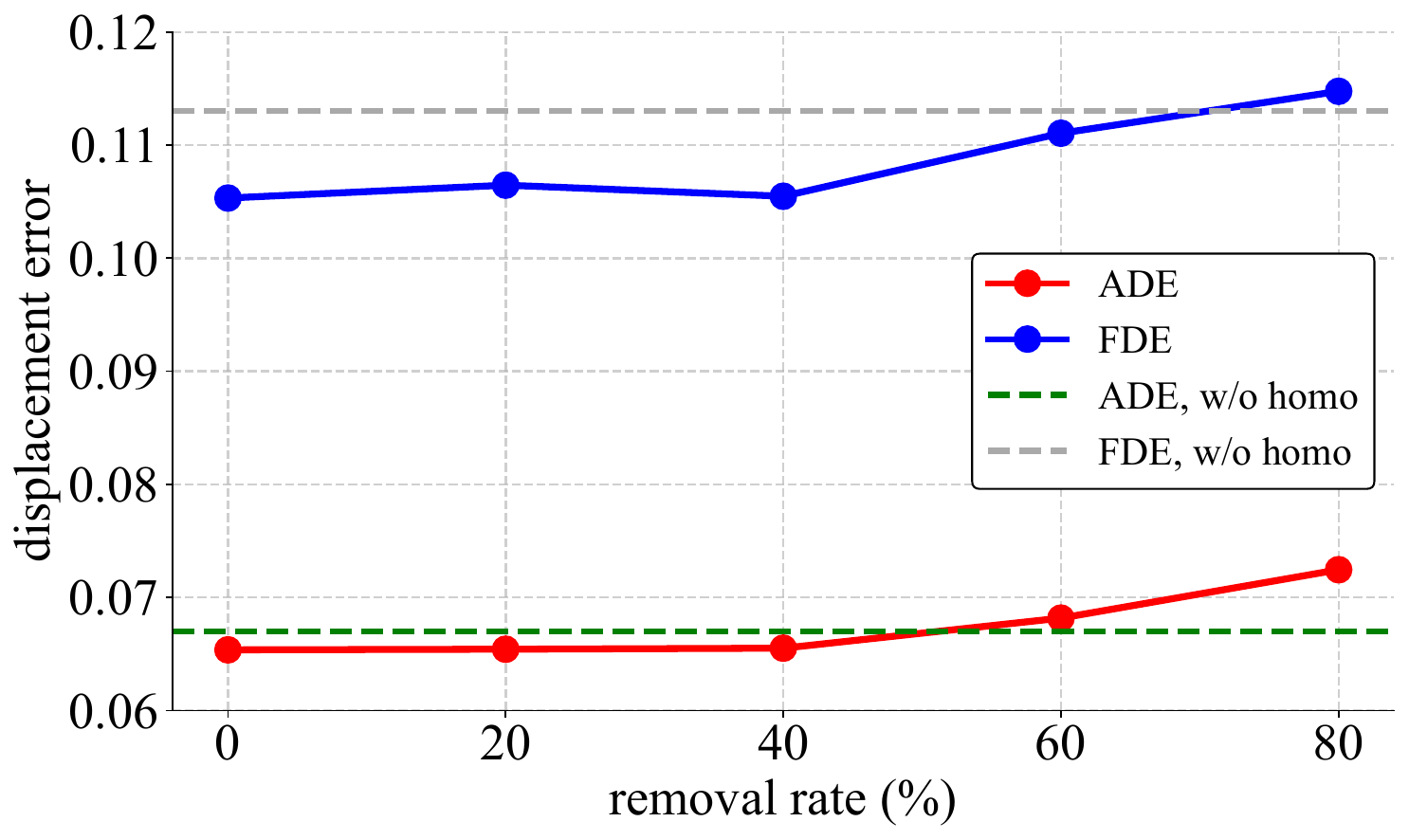}
        \vspace{-0.25cm}
        \caption{\myblue{Keypoint removal (seen)}}
        \label{fig:rm_kp_seen}
    \end{subfigure}
    \begin{subfigure}[t]{1\linewidth}
        \centering
        \includegraphics[width=0.9\linewidth]{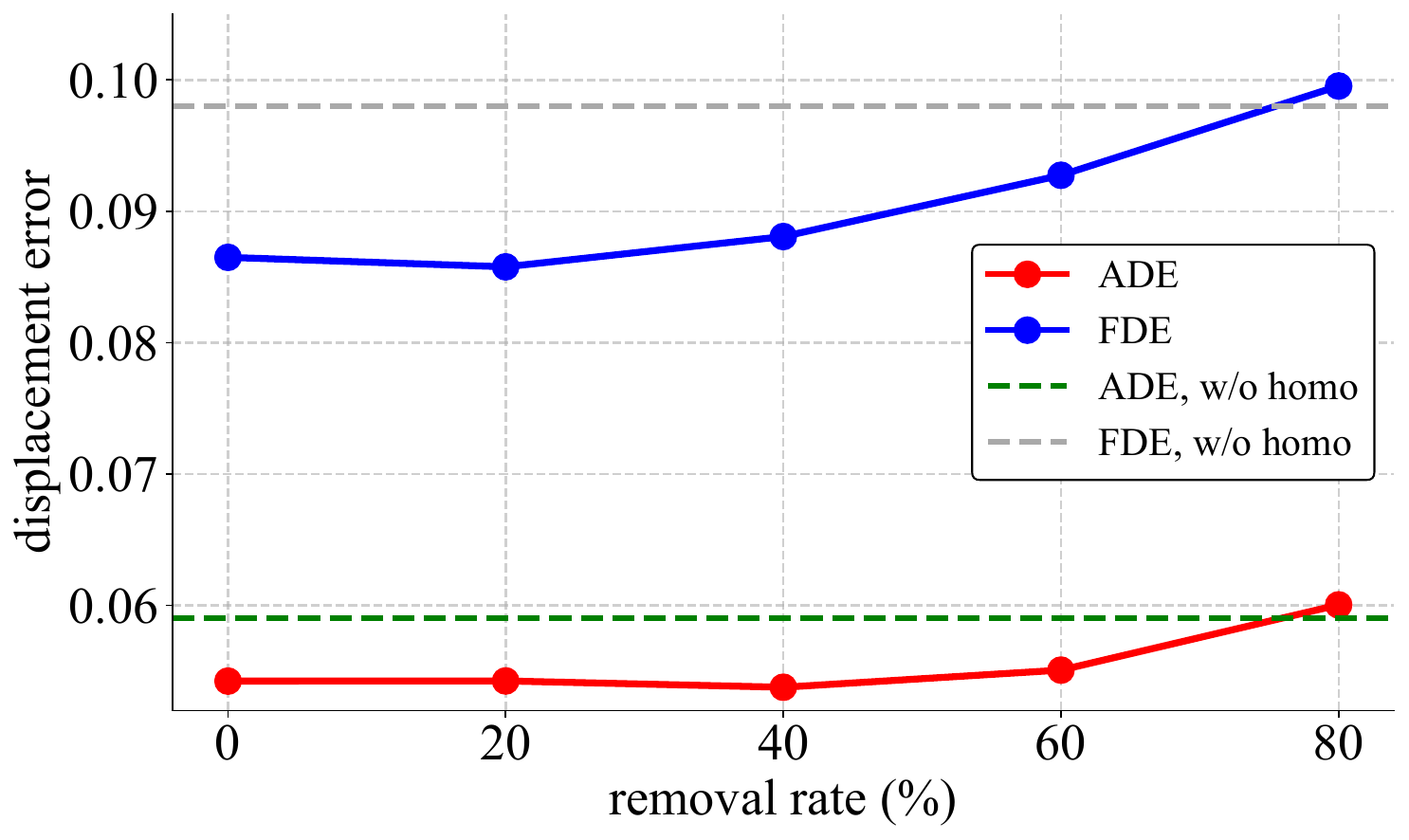}
        \vspace{-0.25cm}
        \caption{\myblue{Keypoint removal (unseen)}}
        \label{fig:rm_kp_unseen}
    \end{subfigure}
    \vspace{-0.25cm}
    \caption{\myblue{Performance changes with keypoint removal that simulates low-texture and dynamic scenes. We report MADiff's prediction errors on EgoPAT3D-DT. The reference performance of MADiff without using camera homography is represented by dashed lines.}}
    \vspace{-0.6cm}
    \label{fig:rm_kp}
\end{figure}

\begin{figure}[t]
    \centering
    \begin{subfigure}[t]{1\linewidth}
        \centering
        \includegraphics[width=0.9\linewidth]{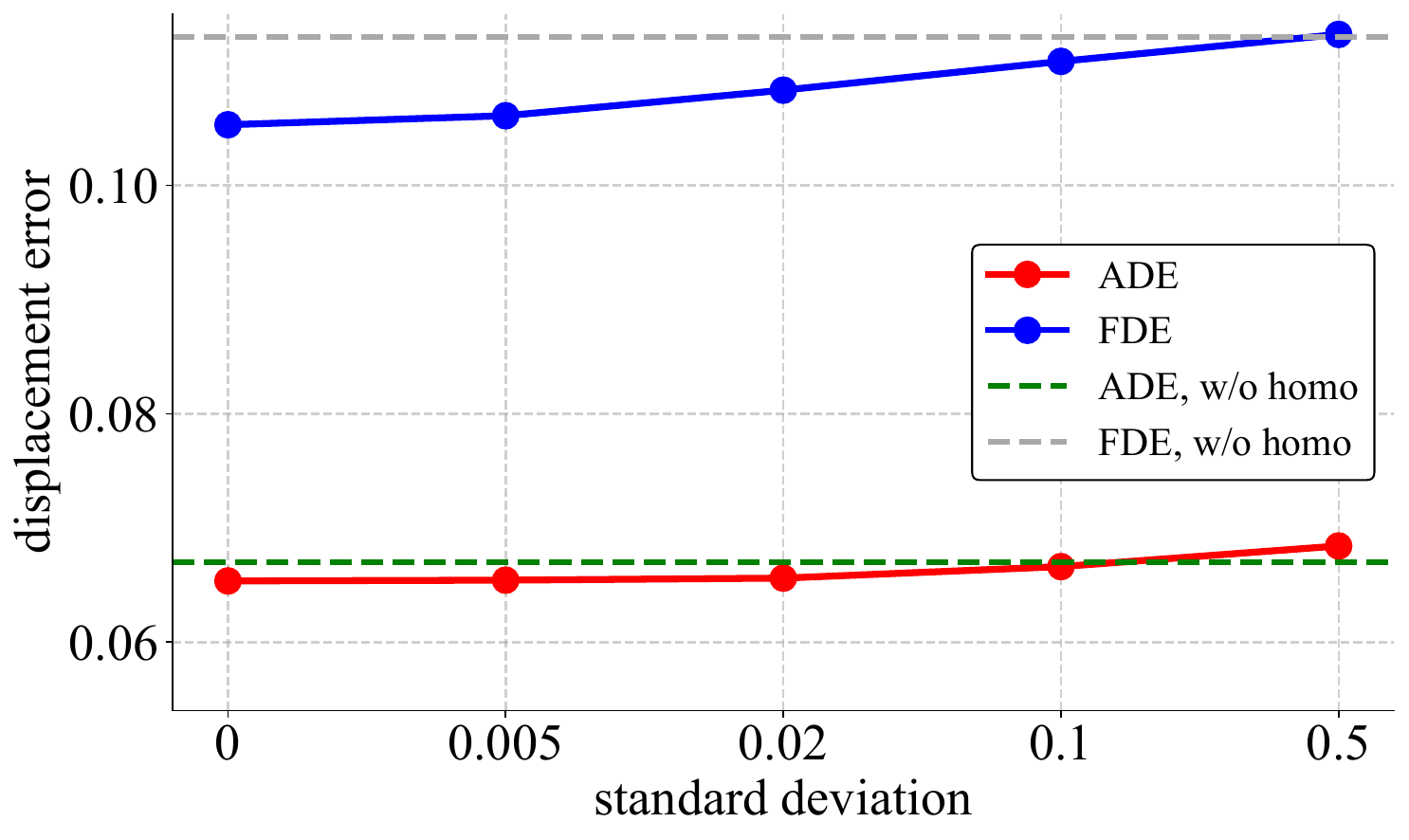}
        \vspace{-0.25cm}
        \caption{\myblue{Gaussian noise injection (seen)}}
        \label{fig:gaussians_seen}
    \end{subfigure}
    \begin{subfigure}[t]{1\linewidth}
        \centering
        \includegraphics[width=0.9\linewidth]{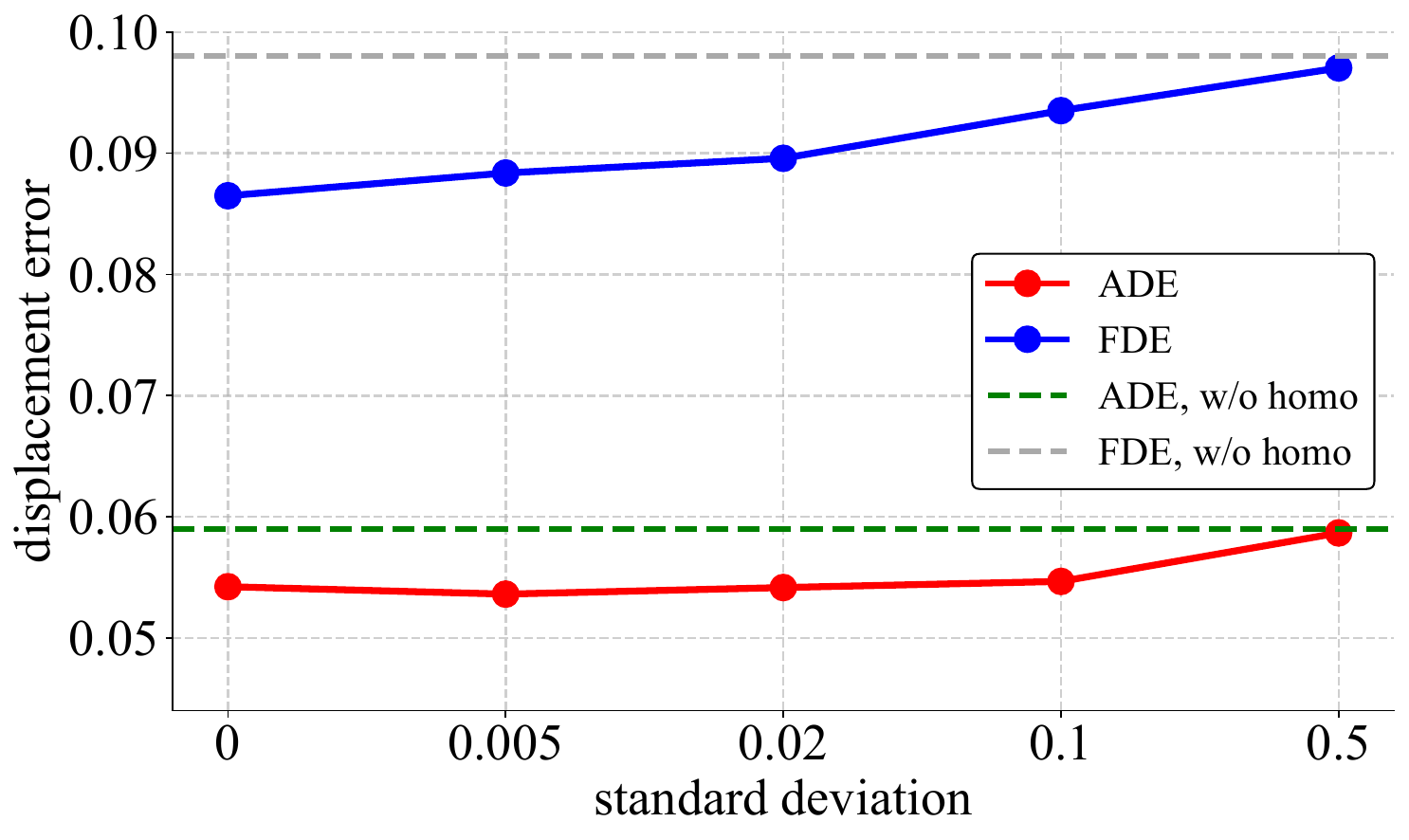}
        \vspace{-0.25cm}
        \caption{\myblue{Gaussian noise injection (unseen)}}
        \label{fig:gaussians_unseen}
    \end{subfigure}
    \vspace{-0.25cm}
    \caption{\myblue{Performance changes with Gaussian noise injection that simulates low-texture and dynamic scenes. We report MADiff's prediction errors on EgoPAT3D-DT. The reference performance of MADiff without using camera homography is represented by dashed lines.}}
    \vspace{-0.3cm}
    \label{fig:gaussians}
\end{figure}

\subsection{Ablation Study on the Number of Mamba Blocks and Diffusion Steps}
\label{sec:abla_mamba_diff}

We conduct the ablation on the number of Mamba blocks with EgoPAT3D-DT. We evaluate $\{0, 2, 4, 6, 8, 10\}$ Mamba blocks in Fig.~\ref{fig:ablation_on_mamba_blocks}. The errors at 0 (MLP) represent the HTP performance of the baseline removing the state transition of SSM in MADiff, which is equivalent to an MLP-based diffusion model. \myblue{The counterpart at 0 (trans) corresponds to the performance of the baseline Diff-IP2D \cite{ma2024diff} that uses denoising transformer rather than Mamba, following the DiT paradigm. Our proposed Mamba diffusion models significantly outperform the MLP- and transformer-based baselines. Although the DiT paradigm of 0 (trans) results in lower prediction errors compared to 0 (MLP), it still struggles to capture the temporal causal
transitions in hand motion states. It cannot well incorporate egomotion features into temporal sequence modeling by the unexplainable attention mechanism, which do not naturally align with the egocentric perspective of hand movements (i.e., how the egomotion affects the hand position change between time steps). This misalignment leads to suboptimal integration of temporal cues.}
MADiff with 4 and 6 Mamba blocks have similar predictive capabilities, while the prediction performance slightly drops when the number of Mamba blocks increases to 8. The reason could be that more Mamba blocks require more data for optimization, and the model with 8 Mamba blocks tends to overfit to our training set.

\myblue{In addition, we conduct an ablation study to evaluate the impact of varying the number of diffusion steps $\{100, 200, 300, 400, 500, 1000, 1500, 2000\}$ in MADiff on EgoPAT3D-DT (EP) and H2O-PT (H2O). As shown in Fig.~\ref{fig:ablation_on_diffusion_steps}, in the early range (100–500 steps), increasing the diffusion steps generally improves prediction quality, indicating that the model benefits from finer-grained denoising. However, beyond 500 steps, performance exhibits only slight fluctuations. A setting of 1000 steps achieves a balanced trade-off between ADE and FDE, and is therefore adopted as the default in MADiff.}

\myblue{
\subsection{Study on Degenerate Conditions in Egomotion Estimation}
\label{sec:abla_on_degen}
We furthermore demonstrate MADiff's robustness to degenerate conditions in egomotion estimation, considering possible low-texture and dynamic scenes in real-world applications. Concretely, we progressively remove partial matched SIFT keypoints (\{20\%, 40\%, 60\%, 80\%\}) before homography calculation, and also inject zero-mean Gaussian noise with increasingly larger standard deviations (\{0.005, 0.02, 0.1, 0.5\}) into computed homographies. Fig.~\ref{fig:rm_kp} shows that the effect of incorporating camera homography consistently remains positive under 60\% keypoint removal rates in both seen and unseen scenes. Fig.~\ref{fig:gaussians} also shows that MADiff is robust to zero-mean Gaussian noises on the estimated camera homography. The reason could be that even when the estimated homography is not highly precise under degenerate conditions, it can still provide a rough trend of camera egomotion, thereby effectively guiding the state transition in temporal modeling of MDSS in the right direction.}

\subsection{Study on the Effect of Multiple Inputs}
\label{sec:abla_on_inputs}
We present the contributions of different combinations of inputs for MADiff on the EgoPAT3D-DT (seen) and EK55. As shown in Tab.~\ref{tab:ala_on_inputs}, only using past hand waypoints as input cannot semantically understand the hand movement in specific scenes, leading to the worst prediction performance of ADE/FDE on EgoPAT3D-DT and SIM on EK55. Once we exploit the visual prompt as an additional input, ADE, FDE of MADiff prediction drop by 11.4\% and 17.6\% respectively, and SIM increases by 15.6\%. Moreover, after importing the text prompt $\mathtt{hand}$, ADE and FDE further decrease by 7.1\% and 6.3\% respectively on EgoPAT3D-DT. SIM of predicted interaction points is also improved by an additional 4.8\% on EK55. The experimental results validate the effectiveness of semantic features generated by our text-guided grounding model for hand trajectory prediction. 
It is also notable in Tab.~\ref{tab:compare_hand_epic_eg} and Tab.~\ref{tab:compare_hand_egopat_h2o} that MADiff outperforms OCT \cite{liu2022joint} and Diff-IP2D \cite{ma2024diff} which require devised global/hand/object features as inputs and are both supervised by additional affordance labels.
We thus argue that the foundation model can capture the relationships between hands and scenarios, avoiding the need for additional task-specific features and affordance labels in the hand trajectory prediction task.

\begin{table}[t]
\small
\setlength{\tabcolsep}{3.5pt}
\renewcommand\arraystretch{1.2}
\center
\caption{Ablation study on multiple inputs, where the scores for the best performance are bolded.}
\vspace{-0.2cm}
\begin{tabular}{ccccc|cc|c}
\toprule
\multicolumn{5}{c|}{Input}     & \multirow{2}{*}{ADE\,$\downarrow$} & \multirow{2}{*}{FDE\,$\downarrow$} & \multirow{2}{*}{SIM\,$\uparrow$} \\ \cline{1-5}
waypoints & visual & $\mathtt{arm}$ & $\mathtt{body}$ & $\mathtt{hand}$  &                       &         &             \\ \midrule
\ding{51}      &       &     &    &      &  0.079                    &  0.136         & 0.199          \\
 \ding{51}             & \ding{51}    &    &   &      & 0.070                     & 0.112         &   0.230         \\
\ding{51}            & \ding{51}    & \ding{51}  &  &   & 0.068                     & 0.109           & 0.236     \\ 
\ding{51}            & \ding{51}    & & \ding{51}   &   & 0.069                     & 0.110  & 0.232  \\ 
\rowcolor{lightgray}
\ding{51}            & \ding{51}    &  &  & \ding{51}  & \textbf{0.065}                     & \textbf{0.105}         &\textbf{0.241}           \\ \bottomrule
\end{tabular}
\label{tab:ala_on_inputs}
\vspace{-0.2cm}
\end{table}

\begin{figure}[t]
    \centering
    \begin{subfigure}[b]{0.32\linewidth}
        \centering
        \includegraphics[width=\linewidth]{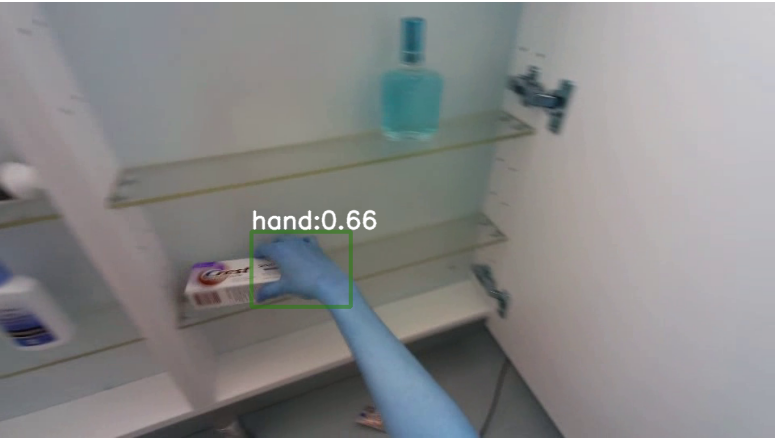}
        \caption{$\mathtt{hand}$}
        \label{fig:hand_prompt}
    \end{subfigure}
    \hfill
    \begin{subfigure}[b]{0.32\linewidth}
        \centering
        \includegraphics[width=\linewidth]{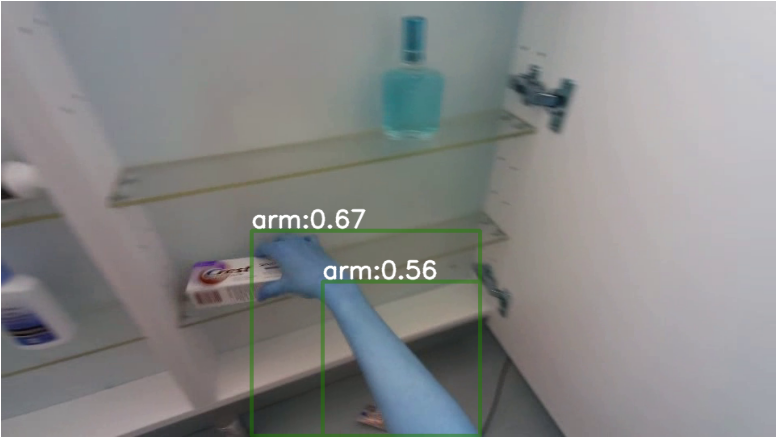}
        \caption{$\mathtt{arm}$}
        \label{fig:arm_prompt}
    \end{subfigure}
    \hfill
    \begin{subfigure}[b]{0.32\linewidth}
        \centering
        \includegraphics[width=\linewidth]{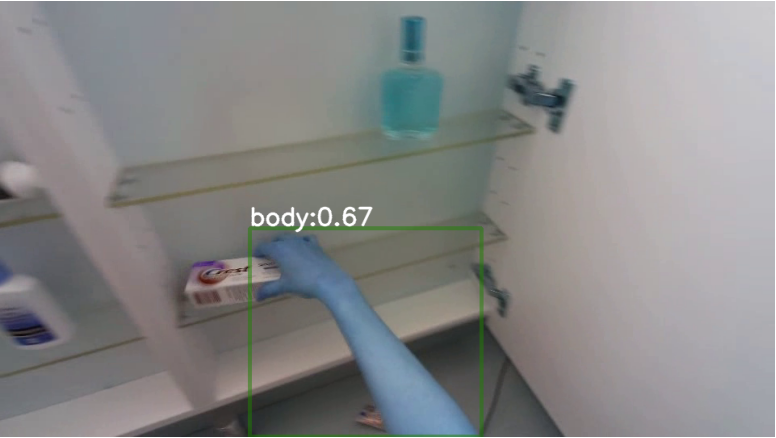}
        \caption{$\mathtt{body}$}
        \label{fig:body_prompt}
    \end{subfigure}
    \vspace{-0.25cm}
    \caption{Visual grounding examples with different input text prompts.}
    \vspace{-0.2cm}
    \label{fig:multiple_text_prompt}
\end{figure}

We also present the effectiveness of two additional text prompts, $\mathtt{arm}$ and $\mathtt{body}$ except for $\mathtt{hand}$. Fig.~\ref{fig:multiple_text_prompt} illustrates respective visual grounding patterns from these text prompts, which also lead to different semantic features. As shown in Tab.~\ref{tab:ala_on_inputs}, the text prompt $\mathtt{hand}$ leads to better prediction than $\mathtt{arm}$ and $\mathtt{body}$ on both two datasets, and the reason could be that a model that intentionally concentrates more on hands has a better understanding of hand movement pattern. Over-focusing on the arm part may cause interference in the model optimization since the closer the arm is to the body, the weaker the correlation between the arm’s swing and the hand trajectories becomes.

\subsection{GLIP vs. Other Backbones in MADiff}
\label{sec:clip_resnet}

In this experiment, we build two other baselines with CLIP \cite{radford2021learning} and ResNet-18 \cite{he2016deep} as image backbones. \myblue{CLIP is task-agnostically transferred to our model to embed each image to a feature vector, which is further fused with the trajectory feature by MLP as diffusion latents.} In contrast, we integrate ResNet-18 into MADiff and train it from scratch. Note that both baselines lack a text prompt compared to GLIP of MADiff. The experiment is conducted on EgoPAT3D-DT and the results in Tab.~\ref{tab:ala_on_glip} show that the utilization of GLIP in MADiff presents the best prediction performance in both previously seen and unseen scenes. The pretrained CLIP cannot generate task-specific semantic features due to a lack of text guidance, and ResNet-18 trained from scratch suffers from overfitting to the previously visited scenarios. 

\begin{table}[t]
\small
\setlength{\tabcolsep}{6.6pt}
\center
\caption{Ablation study on visual backbones, where the scores for the best performance are bolded. There is no text prompt for the baselines with CLIP and ResNet-18.}
\vspace{-0.2cm}
\begin{tabular}{l|cc|cc}
\toprule
\multicolumn{1}{l|}{\multirow{2}{*}{Approach}}   & \multicolumn{2}{c|}{Seen} & \multicolumn{2}{c}{Unseen}   \\ \cmidrule{2-5} 
\multicolumn{1}{c|}{}                                                                               & ADE\,$\downarrow$    & FDE\,$\downarrow$ & ADE\,$\downarrow$    & FDE\,$\downarrow$    \\ \cmidrule{1-5}    
CLIP \cite{radford2021learning}   & 0.068    & 0.108  & 0.055        & 0.089            \\
ResNet-18 \cite{he2016deep}    & 0.068    & 0.106  & 0.061        & 0.092            \\
\rowcolor{lightgray}
GLIP \cite{Li_2022_CVPR} (adopted)  & \textbf{0.065}       & \textbf{0.105}     &  \textbf{0.054}    & \textbf{0.086}   \\ \bottomrule
\end{tabular}
\label{tab:ala_on_glip}
\vspace{-0.5cm}
\end{table}

\begin{figure*}[t]
    \centering
    \begin{subfigure}[b]{0.9\linewidth}
        \centering
        \includegraphics[width=\linewidth]{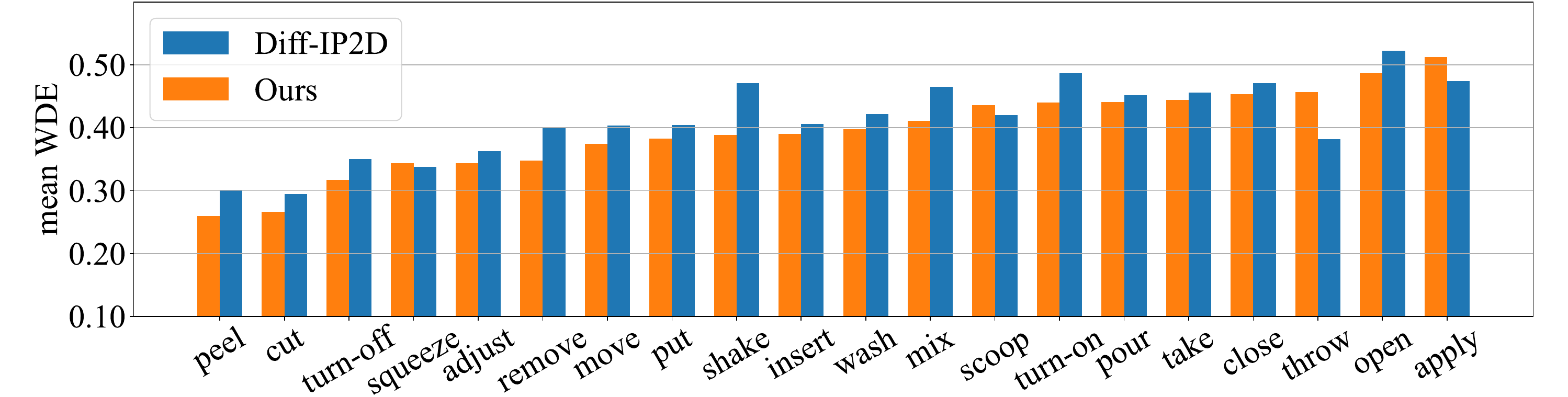}
        \vspace{-0.65cm}
        \caption{Mean WDE vs. action verbs.}
        \label{fig:action_wde}
    \end{subfigure}
    \vfill
    \vspace{0.2cm}
    \begin{subfigure}[b]{0.9\linewidth}
        \centering
        \includegraphics[width=\linewidth]{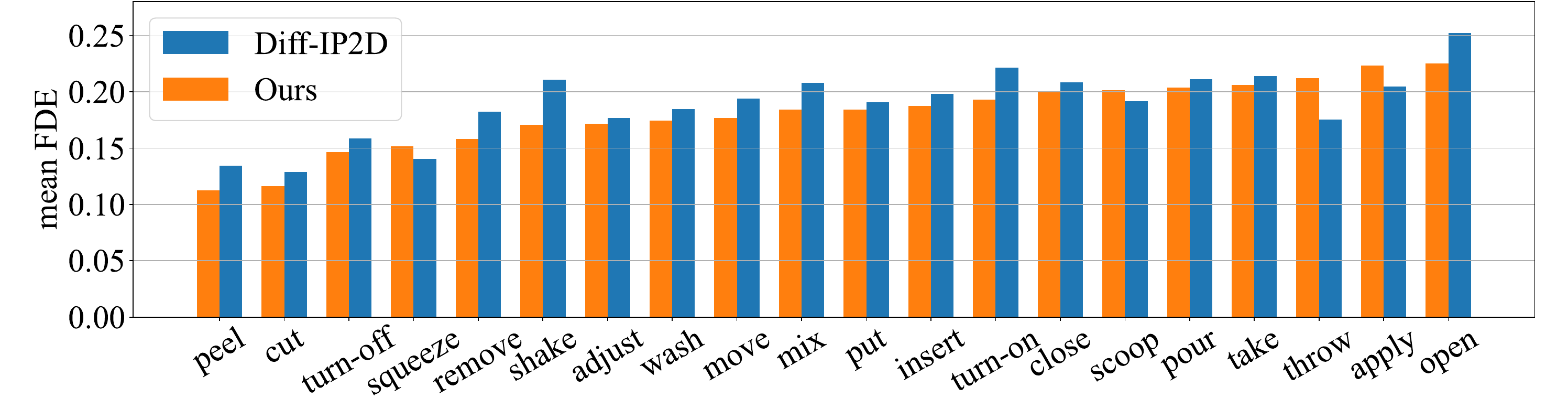}
        \vspace{-0.65cm}
        \caption{Mean FDE vs. action verbs.}
        \label{fig:action_fde}
    \end{subfigure}
    \vspace{-0.25cm}
    \caption{Mean WDE and FDE for predicted hand trajectories belonging to different actions annotated in EK100. We calculate mean errors within each action verb category, e.g., the verb \textit{put} includes \textit{put-down}, \textit{put-from}, \textit{put-of}, \textit{put-on}, and so on. We arrange the top 20 most frequently occurring verb labels from left to right in ascending order of displacement errors from MADiff predictions.}
    \vspace{-0.3cm}
    \label{fig:action_category}
\end{figure*}

\subsection{Displacement Errors on Different Action Verb Categories}

This is the first work to report the correlation between displacement errors and multiple action verb categories in the realm of hand trajectory prediction. The experimental justification is attributed to the fact that how a hand moves in a video clip can be concretely summarized by an action verb, and each verb category basically exhibits potential similarity in its corresponding set of hand trajectories.
As the comparison in Fig.~\ref{fig:action_category}, MADiff shows better prediction performance in most verb categories compared to the baseline Diff-IP2D~\cite{ma2024diff}, exceptionally skilled at predicting fine-grained actions such as \textit{peel}, \textit{cut}, and \textit{shake}. In addition, we also discover that actions that increase the uncertainty of object states (e.g., \textit{turn-on}, \textit{take}, \textit{open}) tend to result in higher trajectory prediction errors compared to their opposite counterparts (e.g., \textit{turn-off}, \textit{put}, \textit{close}). Our proposed MADiff generally outperforms the baseline even though there is high uncertainty in the ultimate state of the active object due to high-level scene understanding and temporal causality capture inherent in our paradigm.

\begin{table}[t] \color{black}
\small
\setlength{\tabcolsep}{3pt}
\center
\caption{\myblue{Text prompt tuning with respect to specific action verbs. We select two verbs, \textit{throw and scoop}, to conduct new text prompts. WDE and FDE presented here are the averages over the examples belonging to the specific verbs. The scores for the best performance are bolded.}}
\vspace{-0.2cm}
\begin{tabular}{l|l|cc}
\toprule
Verb  & Text prompt    & WDE\,$\downarrow$    & FDE\,$\downarrow$    \\ \cmidrule{1-4}    
throw & $\mathtt{hand}$   & 0.457    & 0.212            \\
 &$\mathtt{hand,\,which\,is\,{throwing}}$ (manual)   & \textbf{0.387}     & \textbf{0.180}              \\
 &$\mathtt{hand,\,which\,is\,{throwing}}$ (automated)   & 0.433     & 0.201             \\ \midrule
scoop & $\mathtt{hand}$   & 0.436    & 0.202            \\
& $\mathtt{hand,\,which\,is\,{scooping}}$ (manual)  & \textbf{0.407}       & \textbf{0.190}     \\
& $\mathtt{hand,\,which\,is\,{scooping}}$ (automated)  & 0.428       & 0.199      \\ \bottomrule
\end{tabular}
\label{tab:study_on_verbs}
\vspace{-0.2cm}
\end{table}

\begin{figure}[t]
    \centering
    \begin{subfigure}[b]{0.95\linewidth}
        \centering
        \includegraphics[width=\linewidth]{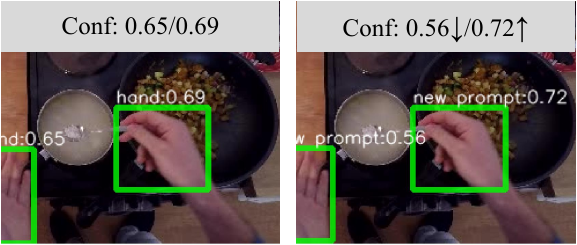}
        \caption{MADiff receives the text prompts of $\mathtt{hand}$ (left) and $\mathtt{hand,\,which\,is\,\textbf{scooping}}$ (right). The annotated verb-noun pair of this example is ``scoop rice''.}
        \vspace{0.2cm}
        \label{fig:scoop_prompt}
    \end{subfigure}
    \begin{subfigure}[b]{0.95\linewidth}
        \centering
        \includegraphics[width=\linewidth]{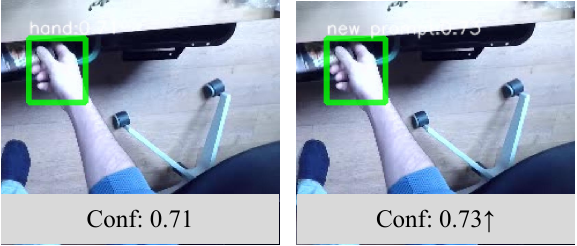}
        \caption{MADiff receives the text prompts of $\mathtt{hand}$ (left) and $\mathtt{hand,\,which\,is\,\textbf{throwing}}$ (right). The annotated verb-noun pair of this example is ``throw tupperware container into the bin''.}
        \vspace{0.2cm}
        \label{fig:throw_prompt}
    \end{subfigure}
    \vspace{-0.4cm}
    \caption{Visual grounding examples with verb-specific prompts. More expressive text prompts lead to changes in confidence.}
    \vspace{-0.7cm}
    \label{fig:multiple_verb_prompt}
\end{figure}

\myblue{Moreover, we also explore how to improve HTP performance for some specific action verbs on EK100. The utilized visual grounding model allows us to adapt verb prompts to generate specific semantic features. Here, we provide both manual and automated text prompt tuning to construct the verb-specific text prompt $\mathtt{hand,\,which\,is\,\{verb\text{-}ing\}}$. The manual implementation follows the GT action annotations for the text prompt of each input video. In contrast, for automated implementation, we use pretrained InAViT~\cite{roy2024interaction} to automatically predict action categories for each video and generate corresponding verb-specific text prompts.
Tab.~\ref{tab:study_on_verbs} indicates that WDE and FDE of the specific verb both decrease significantly if given a more expressive text prompt for both training and testing MADiff. This demonstrates that injecting specific verbs into text prompts helps to generate action-related semantic features, remarkably improving the corresponding HTP accuracy. While the automated implementation is more scalable, it achieves smaller performance gains than the manual implementation due to the inherent prediction errors in action anticipation. Fig.~\ref{fig:multiple_verb_prompt} also implies that the verb-specific prompt encourages the model to focus more on the hand that matches it, according to the changes in confidence.
This experiment overall suggests that MADiff offers a reasonable picture of more flexible HTP solutions than the existing methods, tailored to specific functions in the applications of care robots or other assistive devices.}

\subsection{Inference Time}

We provide the inference time of our proposed MADiff on Epic-Kitchens datasets using the hardware mentioned in Sec.~\ref{sec:madiff_config}. Each prediction by our proposed MADiff costs an average of 0.15\,s, with 0.13\,s for tokenizer and 0.02\,s for the Mamba diffusion process. Since we sample the keyframes in the EK55 and EK100 datasets both with the interval of 0.25\,s, MADiff can predict all the future hand waypoints before the first future keyframe arrives, thus available for online operation. 

\section{Conclusion}
\label{sec:conclusion}

In this paper, we propose a novel hand trajectory prediction method namely MADiff. We first propose using a foundation model to extract high-level semantic features with no need for affordance supervision. Moreover, we design a diffusion model with a devised motion-aware Mamba for denoising. Specifically, the motion-driven selective scan pattern is proposed to fill the motion-related gaps and capture the temporal causality in the continuous denoising step.
We further integrate a continuous-discrete-continuous operation into the diffusion denoising process, combining explicit trajectory iteration with implicit feature iteration.
In addition, we introduce the angle loss and length loss into the training process to facilitate the model capturing directionality and stability better. The experimental results on five publicly available datasets show that our motion-aware Mamba diffusion model MADiff is highly competitive among all the state-of-the-art HTP baselines and the proposed components help improve prediction accuracy effectively. We also present a detailed analysis of MADiff revealing the relationship between prediction errors and action verb categories, providing a critical resource for future research in the field of hand trajectory prediction.

\textbf{Insights and Limitations:} Firstly, our generative paradigm seamlessly integrates Mamba into the denoising diffusion process and bridges autoregressive models and iterative non-autoregressive models, which can serve as a foundation framework for the hand trajectory prediction or other time series forecasting tasks. Secondly, the consideration of egomotion in temporal causality capture provides new insights for diffusion-based techniques in the field of egocentric vision. 
Moreover, our action-relevant analysis opens up a potential direction for future work in the realm of hand trajectory prediction, which is designing distinct prompts specifically for actions of interest.
Despite the encouraging HTP performance, our work still has the following limitations: 1) The specificity of the existing dataset annotations leads to different training and inference setups across different datasets. In the future, we will unify the training and test setups across multiple different datasets. 2) We demonstrate that MADiff can generate good interaction points according to our new evaluation metrics, but it currently cannot actively extract possible affordance maps. We will consider adding a new branch to MADiff, which can achieve affordance prediction for the next active object.

\bibliographystyle{unsrt}
\bibliography{main}

\vspace{-3\baselineskip}
\begin{IEEEbiography}[{\includegraphics[width=1in,height=1.25in,clip,keepaspectratio]{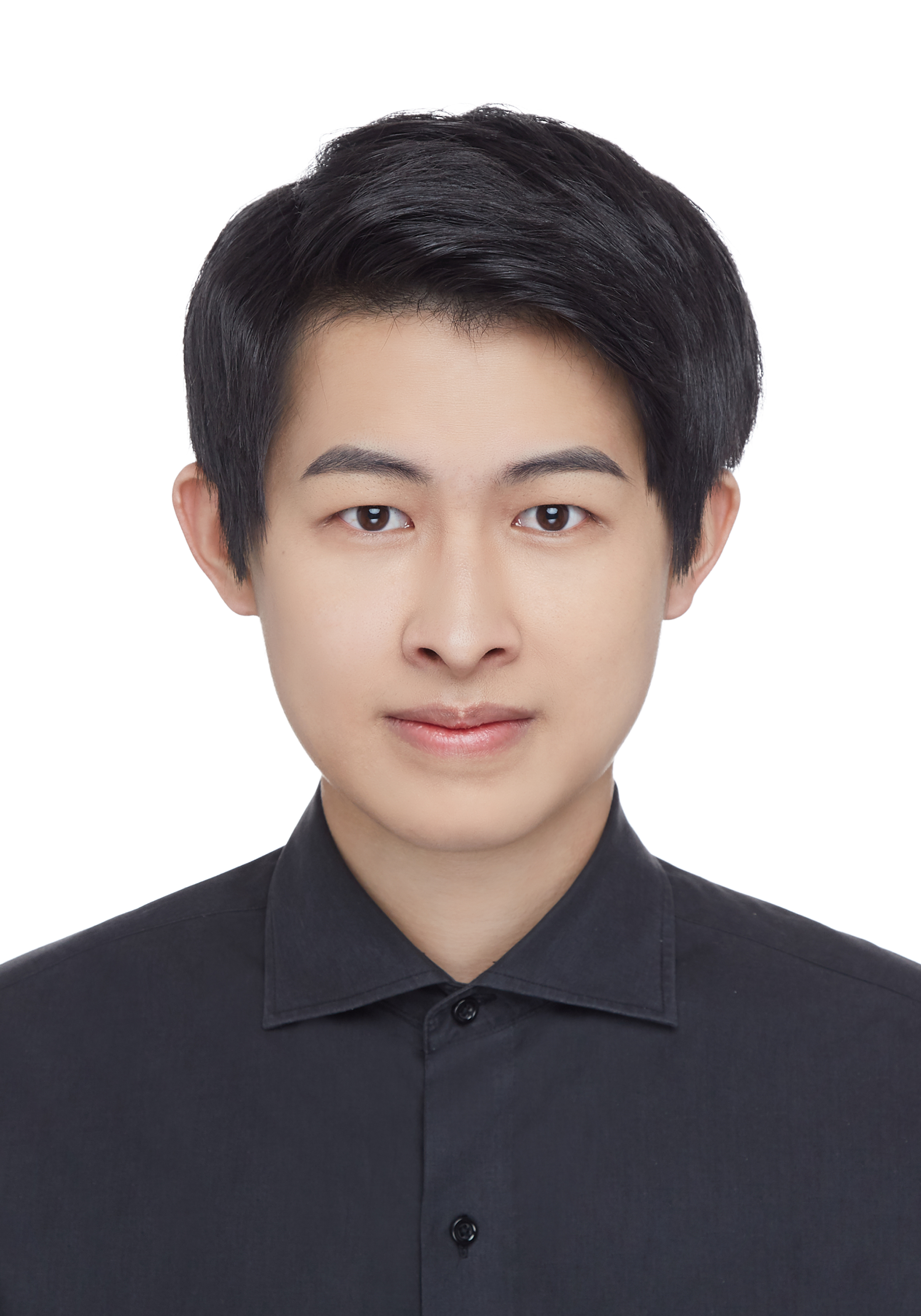}}]
	{Junyi Ma} is a PhD candidate at IRMV Lab, Shanghai Jiao Tong University. He received his Master's degree and Bachelor's degree from the Beijing Institute of Technology, in 2020 and 2023 respectively. He is currently supervised by Prof. Hesheng Wang. His research interests include video analysis and computer vision. He is trying to apply machine learning methods to robotics for enhanced capability of scene understanding.
\end{IEEEbiography}
\vspace{-0.7cm}
\begin{IEEEbiography}[{\includegraphics[width=1in,height=1.25in,clip,keepaspectratio]{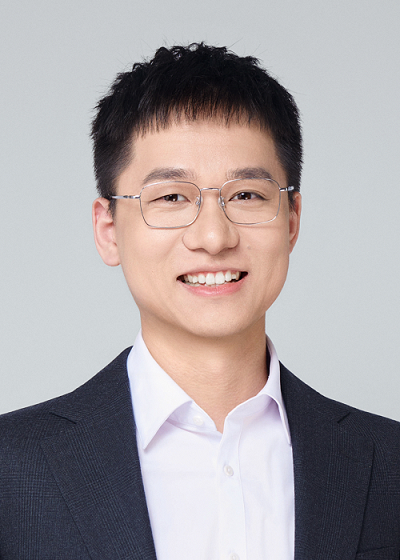}}]
{Xieyuanli Chen} is an Associate Professor at the National University of Defense Technology, China. He received his Ph.D. degree at the Photogrammetry and Robotics Laboratory at the University of Bonn. He received his Master's degree in Robotics in 2017 at the National University of Defense Technology. He received his Bachelor's degree in Electrical Engineering and Automation in 2015 at Hunan University. He serves as Associate Editor for IEEE RA-L, ICRA, and IROS. His research interests are SLAM, localization, mapping, and robot perception. 
\end{IEEEbiography}
\vspace{-0.7cm}
\begin{IEEEbiography}[{\includegraphics[width=1in,height=1.25in,clip,keepaspectratio]{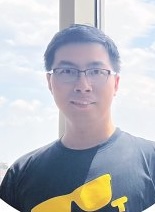}}]
{Wentao Bao} received Ph.D. degree from the Computer Science and Engineering Department at Michigan State University in USA, and BS and MS degrees from Wuhan University in China. His research focuses on open-world computer vision, including video activity recognition, prediction, and understanding in 2D/3D visual space. He has over 20 research works in leading CV/ML journals or conference venues such as CVPR, ICCV, ECCV, ACM MM, IEEE TIP, etc.
\end{IEEEbiography}
\vspace{-0.7cm}
\begin{IEEEbiography}[{\includegraphics[width=1in,height=1.25in,clip,keepaspectratio]{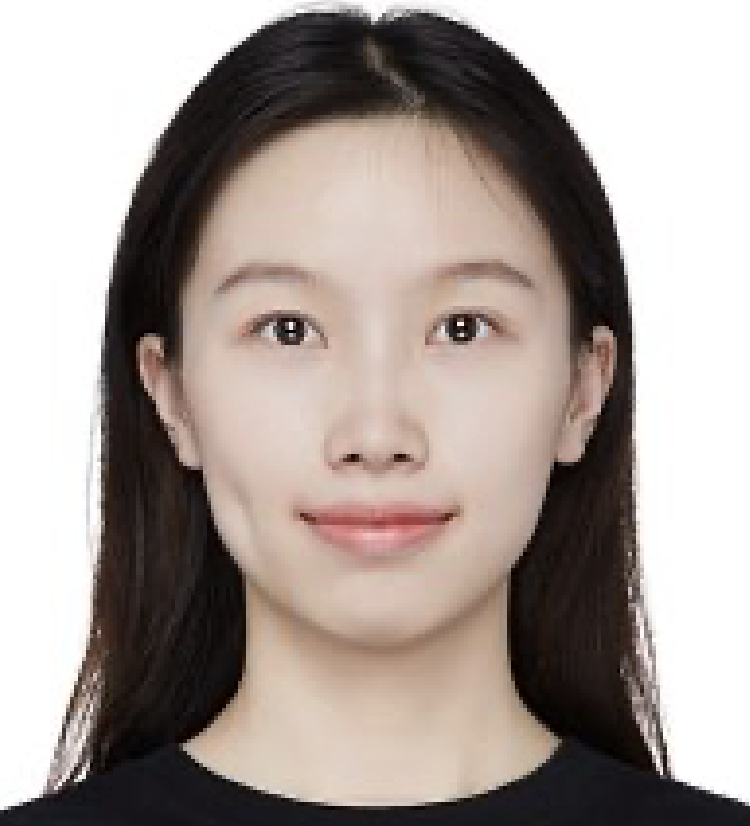}}]
{Jingyi Xu} is working toward the PhD degree from the Department of Electronic Engineering, Shanghai Jiao Tong University. She received her Master's degree and Bachelor's degree from the Beijing Institute of Technology, in 2020 and 2023 respectively. Her research interests are intelligent robots and autonomous driving. 
\end{IEEEbiography}
\vspace{-0.7cm}
\begin{IEEEbiography}[{\includegraphics[width=1in,height=1.25in,clip,keepaspectratio]{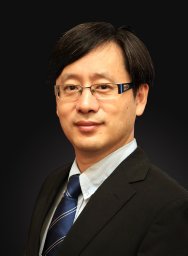}}]
{Hesheng Wang} (Senior Member, IEEE) received the B.Eng. degree in electrical engineering from the Harbin Institute of Technology, Harbin, China, in 2002, and the M.Phil. and Ph.D. degrees in automation and computer-aided engineering from The Chinese University of Hong Kong, Hong Kong, in 2004 and 2007, respectively. He is currently a Distinguished Professor with the School of Automation and Intelligent Sensing, Shanghai Jiao Tong University, Shanghai, China. His current research interests include visual servoing, service robots, soft robots, and computer vision. Prof. Wang serves as a Senior Editor for the IEEE/ASME Transactions on Mechatronics and holds the position of Editor-in-Chief for Robot Learning. He also served as the General Chair of the IEEE/RSJ International Conference on Intelligent Robots and Systems (IROS) in 2025.
\end{IEEEbiography}
\end{document}